\documentclass{article}

\PassOptionsToPackage{numbers, compress}{natbib}
\usepackage[preprint]{neurips_2026}

\usepackage[utf8]{inputenc} % allow utf-8 input
\usepackage[T1]{fontenc}    % use 8-bit T1 fonts
\usepackage{hyperref}       % hyperlinks
\usepackage{url}            % simple URL typesetting
\usepackage{booktabs}       % professional-quality tables
\usepackage{amsfonts}       % blackboard math symbols
\usepackage{nicefrac}       % compact symbols for 1/2, etc.
\usepackage{microtype}      % microtypography
\usepackage{xcolor}         % colors
\usepackage{amsmath} 
\usepackage{cleveref}
\usepackage{graphicx}
\usepackage{colortbl}
\usepackage{enumitem}
\usepackage{nicefrac}
\usepackage{tabularx}  
\usepackage{makecell} 
\usepackage{booktabs}
\usepackage{multirow}
\usepackage{array}
\usepackage{threeparttable}
\usepackage{makecell} 
\usepackage[english]{babel}
\usepackage{blindtext}
\usepackage{capt-of}
\usepackage{placeins}
\usepackage{enumitem}
\usepackage{amssymb}
\usepackage[table]{xcolor}
\usepackage{subcaption}
\usepackage{listings}
\usepackage{mdframed} 
\usepackage{caption}
\usepackage[most]{tcolorbox}
\usepackage{framed}
\usepackage{inconsolata}
\usepackage{wrapfig2}
\usepackage{tikz}
\usepackage{adjustbox}
\usepackage{floatflt}

\newcommand{\pname}[1]{\textcolor{black}{GeoX}}

\title{\pname{}: Mastering Geospatial Reasoning Through Self-Play and Verifiable Rewards}

\author{%
  Kyeongjin Ahn \\
  KAIST\\
  MPI-SP \\
  \And
  Seungeon Lee \\
  MPI-SWS \\
  \AND
  Krishna P. Gummadi \\
  MPI-SWS \\
  \And
  Meeyoung Cha \\
  MPI-SP \\
}

\begin{document}

\maketitle

\begin{abstract}
Geospatial reasoning requires solving image-grounded problems over the complex spatial structure of a scene. 
However, developing this capability is hindered by the cost of annotating a vast and combinatorial question space.
We propose \pname{}, a self-play framework that acquires spatial logic through executable programs that yield verifiable rewards, without relying on large-scale human-curated data
Given a satellite or aerial image, our framework employs a single multimodal policy that proposes spatial problems as executable programs and solves them under three reasoning modes—abduction, deduction, and induction—over spatial primitives and an image understanding tool.
A verifier executes each program to covert a reward signal that jointly optimizes the two roles via reinforcement learning.
\pname{} consistently improves its base VLMs by up to 5.5 points on average, matching or exceeding conventional baselines trained on millions of curated data.
Alongside the proposed method, we release a benchmark for geospatial understanding accumulated through self-play.
\end{abstract}
   
\section{Introduction}
\label{sec:intro}

Interpreting the complex Earth's surface from high-resolution satellite and aerial imagery is a central challenge in remote sensing, with broad implications for urban planning~\citep{ahn2023fine, ahn2025georeg, ahn2025mapping, lee2026generalizable}, environmental monitoring~\citep{yang2024assessing, song2025measuring}, and disaster response~\citep{gupta2019creating, ahn2025generalizable}.
As such imagery proliferates at unprecedented scale, these capabilities are becoming a strategic necessity.
Many practical questions like \emph{``which buildings lie closest to the planned metro station?''} cannot be answered by recognizing \emph{what} is present in an image alone. 
They instead demand a structured understanding of \emph{how} objects interact, including their arrangement, adjacency, and relations across the scene.
Standard tasks such as scene classification, object detection, and image captioning characterize the \emph{contents}, but not its underlying \emph{spatial structure} of a scene. 
\textit{Geospatial reasoning} over this structure is thus a core capability for real-world deployment.

\begin{figure}[t]
    \centering
    \begin{subfigure}[t]{0.38\linewidth}
        \centering
        \includegraphics[width=\linewidth]{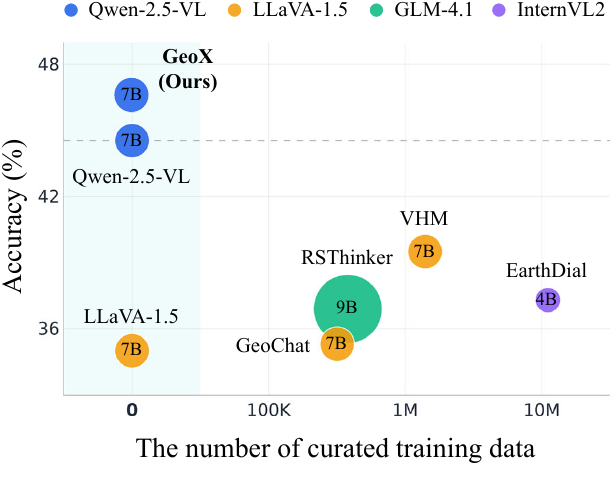}
        \caption{Performance on VQA benchmarks}
        \label{fig:intro-pareto}
    \end{subfigure}
    \hfill
    \begin{subfigure}[t]{0.61\linewidth}
        \centering
        \includegraphics[width=\linewidth]{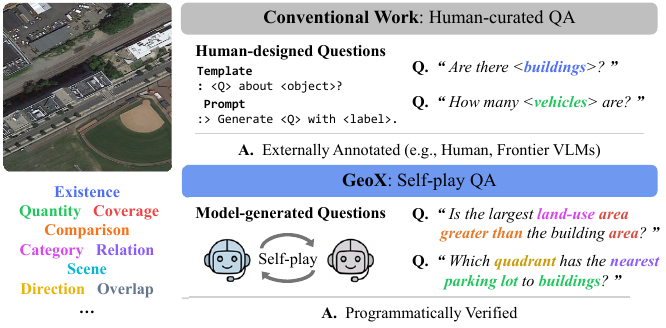}
        \caption{Paradigm comparison}
        \label{fig:intro-paradigm}
    \end{subfigure}
    \caption{%
        \textbf{Standing and motivation.}
        \textbf{(a)} \pname{} outperforms prior VLMs on VQA benchmarks while using zero curated training data;
        \textbf{(b)} Conventional work derives questions from human-designed templates with answers externally annotated, confining them to predefined patterns. Our framework replaces this paradigm with autonomous self-play, where a single model proposes questions whose answers are programmatically verified, broadening them to compositional patterns.
    }
    \label{fig:intro}
\end{figure}

Specialized vision-language models (VLMs) for remote sensing have advanced significantly, but their geospatial understanding is bottlenecked by their supervision regime~\citep{kuckreja2024geochat, hu2025rsgpt, zhang2024earthgpt}. These models depend on fine-tuning over human-curated question–answer pairs, while the space of spatial questions explodes combinatorially with the many objects in each overhead image.
For example, valid questions within an urban image already grow jointly with the target, the reference, and the property among its hundreds of objects, placing comprehensive coverage beyond the reach of manual annotation.
As a result, existing datasets span only a small fraction of this space, leaving the models bounded by the scope of human curation (Figure~\ref{fig:intro-pareto}).

Our key insight is that imagery itself encodes the spatial structure that enables its own interpretation.  
Tasks involving area, centroid, distance, or adjacency reduce to discrete geometric and topological operations whose answers are fixed by the image rather than left to model judgment. Composed into executable programs, these operations admit uniquely determined answers recoverable from the image alone. 
The supervision required for geospatial reasoning is therefore latent in the scene and accessible without external annotation.

We propose a self-play framework \pname{} for geospatial understanding (Figure~\ref{fig:intro-paradigm}).
A single model alternates between two roles, where a \emph{proposer} that constructs an executable problem over the input image and a \emph{solver} that finds its solution.
Each problem specifies an executable program whose execution coverts a verifiable reward that jointly optimizes both roles via reinforcement learning, grounding the learning signal in the spatial structure of the image.
To master a scene rather than merely answer about it, each problem is recast under three reasoning modes: \emph{abduction}, \emph{deduction}, and \emph{induction}.

Our contributions are as follows.
\begin{itemize}[leftmargin=*, itemsep=2pt, topsep=4pt]
    \item \textbf{Self-play with verifiable rewards for geospatial reasoning.} We adapt the self-play paradigm using proposer and solver to remote sensing imagery by routing perception into executable programs composed over a spatial primitives and an image understanding tool. A verifier turns each problem into a verifiable reward signal by program execution, letting the model internalize the geometric and topological structure of the Earth's surface without curated data.
    \item \textbf{One proposal seen through three modes of reasoning.} Each problem is recast under \emph{abduction}, \emph{deduction}, and \emph{induction}, which mask a different element of the problem and elicit a different mode of interrogation: inferring the cause from observed evidence, predicting outcomes from a known cause, and synthesizing the procedure that connects them. One proposal can afford three complementary views of the same scene.
    \item \textbf{Strongest gains where curated supervision falls short.} Across remote sensing benchmarks for visual question answering and object counting, our method  matches or surpasses prior specialized VLMs, with the largest gains on counting and spatial relation tasks, precisely where the cost of curation becomes prohibitive.
    \item \textbf{A self-grown benchmark verified by program execution.} From the self-play process, we automatically accumulate problems with execution-verified answers, ranging from simple to compositional relations. The resulting suite broadens existing remote sensing benchmarks, which skew toward presence detection and scene classification.
\end{itemize}

This framework is a path toward learning the spatial physics of remote sensing imagery. By grounding optimization in executable programs, the framework moves beyond rote prediction toward structural representations that capture inherent properties of the physical world. This aligns with emerging efforts to develop methods that acquire physical laws via self-improving. 
  
\section{The \pname{} paradigm}

\begin{figure}[t]
    \centering
    \hspace{-3mm}
    \includegraphics[width=1.0\linewidth]{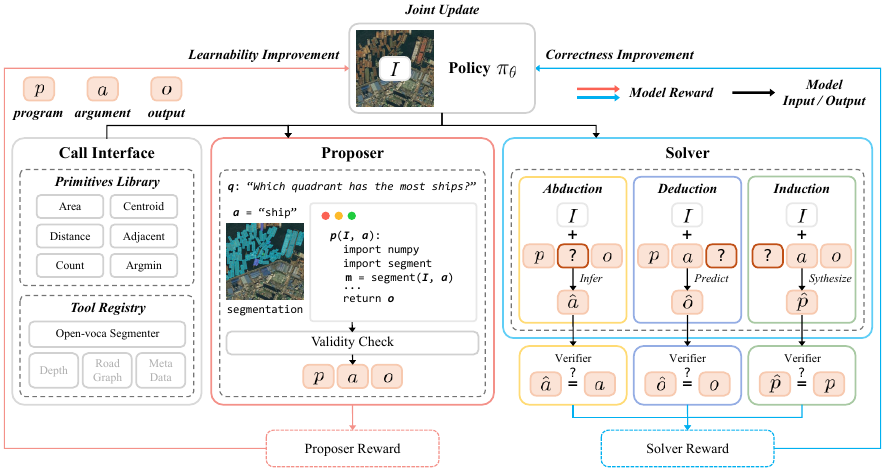}
    \caption{%
        \textbf{Method overview.}
        A single multimodal policy $\pi_\theta$ alternates between a \emph{proposer} (middle) and a \emph{solver} (right) that share a \emph{call interface} (left) of spatial primitives $\mathcal{F}$ and tools $\mathcal{T}$ (instantiated here with an open-vocabulary segmenter; greyed entries left for future work).
        Conditioned on image $I$, the proposer construct a problem by composing calls into an executable program $p$ paired with an argument $a$, forming $(p, a, o)$ with $o = p(a;\, I)$.
        The solver then recovers a masked element of the problem under three reasoning modes: \emph{abduction} infers $\hat{a}$, \emph{deduction} predicts $\hat{o}$, and \emph{induction} synthesizes $\hat{p}$.
        The verifier calculates a \textbf{proposer reward} for problem learnability and a \textbf{solver reward} for answer correctness, which jointly update $\pi_\theta$.
    }
    \label{fig:method}
\end{figure}

\subsection{Preliminaries}
\label{sec:prelim}

We establish the core mechanisms of reinforcement learning through self-play and verifiable rewards.
Due to space constraints, a comprehensive review of additional relevant literature is provided in Appendix~\ref{app:related}.

\paragraph{Reinforcement Learning with Verifiable Rewards (RLVR).}
In RLVR, a model generates a response $y$ for a question $x$ and scores it against a reference answer $y^\star$ using a verifier via exact-match or program execution.
Formally, it maximizes the expected reward over a dataset of question-answer pairs $\mathcal{D}_{\text{RLVR}} = \{(x_n, y_n^\star)\}_{n=1}^N$:
\begin{equation}
    \mathcal{J}_{\text{RLVR}}(\theta)
    =
    \mathbb{E}_{(x, y^\star) \sim \mathcal{D}_{\text{RLVR}},\; y \sim \pi_\theta(\cdot \mid x)}
    \big[ r(y, y^\star) \big],
    \label{eq:rlvr}
\end{equation}
where $\pi_\theta$ is the policy and $r$ is a verifier reward (e.g., $\mathbb{I}[y = y^\star]$).

This approach bypasses human preference labels and learned reward models, delivering a direct and stable feedback signal. 
Whereas supervised fine-tuning (SFT) teaches a model to imitate every token of a reference response, RLVR constrains it on only the final answer, leaving the intermediate reasoning trajectory free to be explored and refined.

\paragraph{Self-play between proposer and solver}
Self-play enables a model to improve by posing and addressing its own problems. In one such setup, a \textbf{proposer} and a \textbf{solver} share parameters to form a closed learning cycle~\citep{zhao2025absolutezero}. The proposer constructs an executable problem, the solver attempts to solve it, and a verifier executes the program to provide a reward that optimizes both roles. Each problem is represented as a triplet:
\begin{equation}
    (p,\, a,\, o), \qquad o = p(a),
    \label{eq:triplet}
\end{equation}
where $p$ is an executable program, $a$ is its input argument, and $o$ is its output.
We use three distinct reasoning modes by masking different elements of the triplet:
In \emph{abduction}, the solver infers $a$ from $(p, o)$.
In \emph{deduction}, it predicts $o$ from $(p, a)$.
In \emph{induction}, it synthesizes $p$ from input-output examples $\{(a_t, o_t)\}_t$.

Concretely, $p$ is a program that counts ships of a specified type in a geospatial context. Here, \textit{abduction} hypothesizes the type from an observed count ($o=7 \to a=$\,\texttt{cargo}); \textit{deduction} derives the count from a given type ($a=$\,\texttt{cargo} $\to o=7$); \textit{induction} generalizes $p$ from input-output pairs ($\{(\texttt{cargo}, 7), (\texttt{fishing}, 3)\} \to p$). 
One proposal thus affords three complementary perspectives on the same scene, each engaging a particular facet of spatial reasoning.

%%%%%%%%%%%%%%%%%%%%%%%%%%%%%%%%%%%%%%%%%%%%%%%%%%%%%%%%%

\subsection{Problem setting}
\label{sec:method:setup}
We refer to the proposed method as \pname{}, which trains a single multimodal policy $\pi_\theta$ on remote sensing imagery through self-play and verifiable rewards via program execution, as illustrated in Figure~\ref{fig:method}. 
The policy first acts as a \textbf{proposer}, constructing an executable problem that realizes a spatial question over an input image. It then acts as a \textbf{solver}, attempting to find the solution for constructed problems.
Both roles are driven by verifiable rewards: the proposer is rewarded for devising challenging yet learnable problems, while the solver is rewarded for answering correctly under the three reasoning modes.

Let $\mathcal{I} = \{I_m\}_{m=1}^{M}$ be a collection of unlabeled remote sensing images. 
The policy $\pi_\theta$ returns an answer $y$ to a spatial question $q$ on an image $I \in \mathcal{I}$. 
Each question is expressed as an image-grounded executable problem:
\begin{equation}
    (p,\, a,\, o,\, I), \qquad o = p(a;\, I),
    \label{eq:task}
\end{equation}
where $p$ is a program, $a$ is its argument, and $o$ is the output. 
For each image $I$, the proposer generates $(p, a)$, whose execution determines $o$. The solver then completes a masked element of the problem from partial observations, depending on the reasoning mode.

%%%%%%%%%%%%%%%%%%%%%%%%%%%%%%%%%%%%%%%%%%%%%%%%%%%%%%%%%

\paragraph{Proposer.}
We maintain three problem banks $\mathcal{D}_{\text{abd}}$, $\mathcal{D}_{\text{ded}}$, and $\mathcal{D}_{\text{ind}}$ corresponding to each reasoning mode. These banks offer in-context references for the proposer.

In \textit{abduction} and \textit{deduction}, the proposer samples two image batches $\mathcal{B}_{\text{abd}}=\{I_b^{\text{abd}}\}_{b=1}^{B}$ and $\mathcal{B}_{\text{ded}}=\{I_b^{\text{ded}}\}_{b=1}^{B}$.
It constructs one executable problem per image $I$ using $K$ in-context references drawn from $\mathcal{D}_{\text{abd}}$ for abduction or $\mathcal{D}_{\text{ded}}$ for deduction.
Each problem consists of a question $q$, program $p$, and argument $a$, with output $o$ obtained by $p(a; I)$. 
Only valid constructions are used for growing the respective bank, while failed (e.g., syntax or runtime errors) or non-deterministic ones (e.g., randomness, file I/O) are discarded.

In \textit{induction}, the proposer samples $B$ programs from $\mathcal{D}_{\text{abd}} \cup \mathcal{D}_{\text{ded}}$ and generates $N_{\mathrm{io}}$ input-output pairs $\{(a_t, o_t)\}_{t=1}^{N_{\mathrm{io}}}$ for each program using references from $\mathcal{D}_{\text{ind}}$. 
The same grow-and-discard rule is applied.

\paragraph{Solver.}
The solver processes $3B$ total problems for three reasoning modes, where samples from each bank fill any shortfall in valid proposals. For each problem, the solver makes $R$ attempts, yielding $3BR$ outputs per training step. In \textit{abduction}, the solver infers the missing argument $\hat{a}$ from $(I, p, o)$. In \textit{deduction}, it predicts the expected output $\hat{o}$ from $(I, p, a)$. In \textit{induction}, it discovers the underlying program $\hat{p}$ from $(I, \{(a_t, o_t)\}_{t=1}^{N_{\mathrm{io}}})$. The verifier scores each result in a mode-specific manner:
\begin{align}
    \text{Abduction:}\quad 
    & \pi^{\text{solve}}_\theta\bigl(\hat{a} \mid I,\, p,\, o\bigr), 
    && o \stackrel{?}{=} p(\hat{a}; I), 
    \label{eq:our-abd}\\[2pt]
    \text{Deduction:}\quad 
    & \pi^{\text{solve}}_\theta\bigl(\hat{o} \mid I,\, p,\, a\bigr), 
    && \hat{o} \stackrel{?}{=} p(a; I), 
    \label{eq:our-ded}\\[2pt]
    \text{Induction:}\quad 
    & \pi^{\text{solve}}_\theta\bigl(\hat{p} \mid I,\, \{(a_t, o_t)\}_{t \in \mathcal{V}}\bigr), 
    && \frac{1}{|\mathcal{U}|}\sum_{t \in \mathcal{U}} \mathbb{I}\bigl[o_t \stackrel{?}{=} \hat{p}(a_t; I)\bigr].
    \label{eq:our-ind}
\end{align}
Here, for induction, the $N_{\mathrm{io}}$ input-output pairs are split into a visible subset $\mathcal{V}$ and a held-out subset $\mathcal{U}$, where $\mathcal{V} \cup \mathcal{U} = \{1,\dots,N_{\mathrm{io}}\}$. 
This requires the solver to reconstruct the program rather than memorize seen pairs.

\paragraph{Seed initialization.}
To initialize this process, we define a simple hand-written seed template that calls a segmentation function to check the presence of an object (e.g., \texttt{"Is there a building?"}), as shown in Figure~\ref{fig:seed}.

%\small
\begin{wrapfigure}[19]{r}{0.43\linewidth}
\vspace{-1.8em}
\centering
\begin{tcolorbox}[
    colback=white,
    colframe=black!15,
    boxrule=0.6pt,
    arc=2pt,
    left=10pt, right=10pt, top=8pt, bottom=8pt,
    boxsep=0pt,
    fontupper=\small,
]
{\scshape\bfseries\footnotesize Seed Program}\par
\vspace{2pt}
{\color{black!15}\hrule height 0.4pt}
\vspace{8pt}

{\scriptsize\color{black!55}QUESTION}~{\scriptsize\color{black!40}$q$}\\
\texttt{\small "Is there a building?"}

\vspace{6pt}

{\scriptsize\color{black!55}ARGUMENT}~{\scriptsize\color{black!40}$a$}\\
\texttt{\small "building"}

\vspace{8pt}

{\scriptsize\color{black!55}PROGRAM}~{\scriptsize\color{black!40}$p$}
\vspace{2pt}

\begin{tcolorbox}[
    colback=black!4,
    colframe=black!4,
    boxrule=0pt,
    arc=2pt,
    left=4pt, right=4pt, top=3pt, bottom=3pt,
    boxsep=0pt,
]
\begin{lstlisting}[
    language=Python,
    basicstyle=\ttfamily\footnotesize,
    keywordstyle=\color{blue!55!black},
    stringstyle=\color{red!55!black},
    showstringspaces=false,
    frame=none,
    aboveskip=0pt, belowskip=0pt,
    xleftmargin=0pt,
]
def f(I, a):
    from tools import segment
    import numpy as np
    mask = segment(I, a)
    return bool(np.any(mask))
\end{lstlisting}
\end{tcolorbox}

\vspace{8pt}

{\scriptsize\color{black!55}OUTPUT}~{\scriptsize\color{black!40}$o$}\\
\texttt{\small True}
\end{tcolorbox}
\caption{\textbf{Seed problem.} A segmenter call with the phrase \texttt{"building"}, followed by a presence check on the returned mask.}
\label{fig:seed}
% \vspace{-0.8em} 
\end{wrapfigure}
The template is instantiated by pairing random object phrases with images, populating each bank with $N_{\mathrm{seed}}$ problems. 
From this warm start, the proposer moves toward increasingly compositional problems, such as comparing areas across object categories, without further human intervention.

%%%%%%%%%%%%%%%%%%%%%%%%%%%%%%%%%%%%%%%%%%%%%%%%%%%%%%%%%

\subsection{Programs over image-grounded interface}
\label{sec:method:programs}
The executor exposes two sets of callables: a primitives library $\mathcal{F}$ and a tool registry $\mathcal{T}$, with $\mathcal{E} = \mathcal{F} \cup \mathcal{T}$ as the full call interface.
The primitives library $\mathcal{F}$ contains geometric operators (e.g., \texttt{area}, \texttt{centroid}, \texttt{orientation}) and topological operators (e.g., \texttt{distance}, \texttt{adjacent}, \texttt{overlaps}), along with aggregation operators (e.g., \texttt{count}, \texttt{argmin}, \texttt{compare}), implemented on top of NumPy, SciPy, and scikit-image.
The tool registry $\mathcal{T}$ contains specialized modules for extracting structured information from the image.
A spatial program composes callables from $\mathcal{E}$, allowing primitives and tools to be interleaved as needed.

\paragraph{A minimal tool registry.}
For the experiments in this paper, we equip $\mathcal{T}$ with a single tool: an open-vocabulary segmenter that maps an image paired with a natural-language phrase (e.g., \texttt{"red-roofed building"}) to a set of segmentation masks.
This minimal choice is motivated by three considerations.
(1) \emph{Spatial sufficiency}. Many spatial tasks considered in this work can be computed directly from segmentation masks using geometric, topological, and aggregation operators.
(2) \emph{Open-vocabulary flexibility}. Open-vocabulary segmenter accepts arbitrary phrases generated by the policy, so a single tool can support diverse spatial questions without committing to a fixed class vocabulary.
(3) \emph{Methodological isolation}. The single-tool usage isolates self-play's contribution from that of a richer toolset, establishing a clean baseline against which future extensions can be evaluated.

\paragraph{Intermediate values are hidden from the policy.}
The policy $\pi_\theta$ accesses only $I$ and the available calls in $\mathcal{F}$ and $\mathcal{T}$.
All intermediate values produced during execution remain internal to the executor and inaccessible to the policy.
For example, when the policy issues a segmenter call with the phrase \texttt{"solar panel array"}, neither the resulting mask array nor any values derived from it in subsequent calls such as \texttt{count} or \texttt{area} are exposed to the policy.
This design prevents the policy from exploiting execution traces as shortcuts, forcing it to reason from the image and program specification instead.

%%%%%%%%%%%%%%%%%%%%%%%%%%%%%%%%%%%%%%%%%%%%%%%%%%%%%%%%%

\subsection{Reward design}
\label{sec:method:reward}
The proposer and solver are optimized jointly through a unified policy $\pi_\theta$ with mutually informative rewards.

\paragraph{Solver reward.}
The solver is rewarded for the \emph{correctness} of its answers by a verifier.
For abduction and induction, where the output is fully determined by program execution, the verifier is reduced to an execution-consistency condition that the returned output under the solver's result exactly matches $o$:
\begin{align}
    r^{\text{solve}}_{\text{abd}}(\hat{a}) 
        = \mathbb{I}\bigl[p(\hat{a}; I) = o\bigr], \quad
    r^{\text{solve}}_{\text{ind}}(\hat{p}) 
        = \frac{1}{|\mathcal{U}|}
        \sum_{t\in\mathcal{U}} \mathbb{I}\bigl[\hat{p}(a_t; I) = o_t\bigr].
    \label{eq:solver_abd_ind_reward}
\end{align}
For deduction, by contrast, where the output may take various types across problems, the type-aware verifier is adopted in place of a binary exact-match rule:
\begin{equation}
    r^{\text{solve}}_{\text{ded}}(\hat{o}) =
    \begin{cases}
        \max\!\bigl(0,\, 1 - |\hat{o} - o|/\max(|o|, 1)\bigr)
            & \text{numeric}, \\[2pt]
        \mathbb{I}[\mathrm{norm}(\hat{o}) = \mathrm{norm}(o)]
            & \text{string}, \\[2pt]
        \mathbb{I}[\hat{o} = o]
            & \text{boolean}, \\[2pt]
        \mathrm{IoU}(\hat{o}, o)
            & \text{bounding box}.
    \end{cases}
    \label{eq:solver_ded_reward}
\end{equation}

\paragraph{Proposer reward.}
The proposer is rewarded for the \emph{learnability}
\citep{zhao2025absolutezero} of its proposals.  
The learnability of a problem is estimated from $R$ solver rollouts, by averaging their solver rewards:
\begin{equation}
    \bar{r}^{\text{solve}}
    \;=\;
    \frac{1}{R}\sum_{i=1}^{R} r^{\text{solve}}_{(i)} ,
\end{equation}
which then defines the proposer reward:
\begin{equation}
    r^{\text{prop}}
    \;=\;
    \mathbb{I}\!\left[\bar{r}^{\text{solve}} > 0\right]
    \cdot
    \left(1 - \bar{r}^{\text{solve}}\right).
    \label{eq:reward-propose}
\end{equation}
The reward vanishes at both extremes ($\bar{r}^{\text{solve}}\!\in\!\{0,1\}$) and peaks at $\bar{r}^{\text{solve}} = 1/2$.
An adaptive curriculum thus emerges, in which the proposer favors problems at the frontier of the solver's ability, advancing as the solver improves

\paragraph{Joint optimization.}
The shared parameters $\theta$ are trained with a joint objective:
\begin{equation}
    \mathcal{J}^{\text{\pname{}}}(\theta) \;=\;
    \lambda\,\mathcal{J}^{\text{prop}}(\theta)
    \;+\;
    \mathcal{J}^{\text{solve}}(\theta),
    \label{eq:rs-zero-total}
\end{equation}
where $\mathcal{J}^{\text{prop}}$ and $\mathcal{J}^{\text{solve}}$ are the expected proposer and solver rewards under their problem distributions. We set $\lambda = 1$ in all experiments, weighting the two roles equally.

\section{Experiments}
\label{sec:exp}

\subsection{Evaluation setup}
\label{sec:exp:setup}

\noindent
\textbf{Datasets.}
Geospatial understanding is evaluated on three VQA benchmarks:
RSVQA-HR~\citep{lobry2020rsvqa}, EarthVQA~\citep{wang2024earthvqa}, and
GEOBench-VLM~\citep{danish2025geobench}. 
From GEOBench-VLM, we use the $17$ out of $31$ tasks that target spatial understanding.

\noindent
\textbf{Baseline models.}
We compare our method against two groups of baselines.
\emph{(i) Base}: general-purpose VLMs evaluated zero-shot, namely Qwen-2.5-VL-7B-Instruct~\citep{wu2025qwen} (Qwen) and LLaVA-1.5-7B~\citep{liu2024improved} (LLaVA).
\emph{(ii) Conventional}: domain-specific VLMs fine-tuned on human-curated remote sensing data, namely  GeoChat~\citep{kuckreja2024geochat}, VHM~\citep{pang2025vhm}, EarthDial~\citep{soni2025earthdial}, and RSThinker~\citep{liu2026towards}.
%Details of each baseline are provided in Appendix~\ref{app:baselines}.

\noindent
\textbf{Evaluation procedure.}
Performance is measured using accuracy.
Closed-form answers are scored by exact match, while open-ended answers are scored by an LLM-as-judge.
For each question, we generate $N_{\mathrm{eval}} = 32$ responses and aggregate them by majority voting.

% Additional details on datasets, baselines, implementation, and self-play prompts are provided in Appendices~\ref{app:datasets},~\ref{app:baselines},~\ref{app:implementation_details}, and~\ref{app:prompts}, respectively.
Appendices~\ref{app:datasets},~\ref{app:baselines}, and~\ref{app:implementation_details}, respectively.
Further analyses on training dynamics and object counting are presented in Appendix~\ref{app:additional-results}, and full results for Sections~\ref{sec:exp:main} and~\ref{sec:exp:ablation} are reported in Appendix~\ref{app:full-results}.

\colorlet{curatedcol}{gray!12}
\colorlet{gaincol}{cyan!10}

\newcolumntype{M}{>{\centering\arraybackslash}p{1.15cm}}
\newcolumntype{C}{>{\columncolor{curatedcol}\centering\arraybackslash}p{1.15cm}}
\newcolumntype{O}{>{\centering\arraybackslash}p{1.15cm}}

\begin{table}[t]
\centering
\footnotesize
\renewcommand{\arraystretch}{1.2}

\caption{
\textbf{Main results on VQA benchmarks.}
Performance of \pname{}, instantiated from two base VLMs (LLaVA and Qwen), is compared against six baselines on RSVQA-HR, EarthVQA, and the four task categories of GEOBench-VLM.
% , with full per-subtask results provided in Appendix~\ref{app:main-subtask}.
%
For each model, the number of human-curated QA pairs used for fine-tuning is listed; baseline models fine-tuned on fully curated data are grey shaded.
The best result in each row is shown in \textbf{bold}, and our results are highlighted in blue when they surpass the corresponding base (LLaVA or Qwen).
}

\begin{adjustbox}{width=0.95\textwidth}
\begin{tabular}{@{}l M M C C C C O O@{}}
\toprule
& \multicolumn{2}{c}{\textit{Base}}
& \multicolumn{4}{c}{\textit{Conventional}}
& \multicolumn{2}{c}{\textit{Self-play}} \\
\cmidrule(lr){2-3} \cmidrule(lr){4-7} \cmidrule(lr){8-9}
\textbf{Benchmark / Task}
  & LLaVA & Qwen
  & GeoChat & VHM & EarthDial & RSThinker
  & \textbf{\pname{}\textsubscript{L}} & \textbf{\pname{}\textsubscript{Q}} \\
\midrule
\textit{\# Curated}
  & 0 & 0 & 318K & 1.4M & 11.1M & 380K & 0 & 0 \\
\midrule

\multicolumn{9}{@{}l}{\textit{RSVQA-HR}} \\
\addlinespace[2pt]
~~Presence                         & 57.9 & 60.1 & 57.9 & \textbf{72.3} & 60.0 & 55.4 & \cellcolor{gaincol}61.9 & \cellcolor{gaincol}63.2 \\
~~Count                            & 30.3 & 35.2 & 25.5 & 13.5 & 25.9 & 22.1 & \cellcolor{gaincol}30.5 & \cellcolor{gaincol}\textbf{36.6} \\
~~Area                             &  2.7 & 10.3 & \textbf{41.6} & 13.5 & 38.8 & 22.1 & \cellcolor{gaincol}40.3 & \cellcolor{gaincol}16.5 \\
~~Comparison                       & 52.5 & 69.7 & 74.8 & 70.7 & \textbf{78.9} & 68.7 & \cellcolor{gaincol}57.5 & \cellcolor{gaincol}72.0 \\
\midrule
~~\textbf{Average}                 & 35.9 & 43.8 & 50.0 & 42.5 & \textbf{50.9} & 42.1 & \cellcolor{gaincol}47.6 & \cellcolor{gaincol}47.1 \\
\midrule

\multicolumn{9}{@{}l}{\textit{EarthVQA}} \\
\addlinespace[2pt]
~~Basic Judging                    & 78.6 & 78.0 & 66.6 & 79.0 & 76.2 & 69.4 & \cellcolor{gaincol}\textbf{82.2} & \cellcolor{gaincol}78.2 \\
~~Reasoning-based Judging          & 55.8 & \textbf{73.0} & 37.0 & 71.8 & 41.4 & 42.6 & 53.8 & 68.6 \\
~~Reasoning-based Counting         & 24.8 & 23.4 & 16.6 & 12.8 &  5.6 & 15.4 & \cellcolor{gaincol}\textbf{33.8} & \cellcolor{gaincol}26.6 \\
~~Basic Counting                   & 60.6 & 62.4 & 26.0 & 39.6 & 53.0 & 45.8 & \cellcolor{gaincol}\textbf{70.2} & \cellcolor{gaincol}69.4 \\
~~Object Situation Analysis        & 28.2 & 29.4 &  2.2 & 23.2 &  4.0 & 25.2 & \cellcolor{gaincol}\textbf{30.4} & \cellcolor{gaincol}30.2 \\
~~Comprehensive Analysis           & 19.6 & 33.6 &  7.8 & 24.6 & 11.0 & 19.0 & 19.5 & \cellcolor{gaincol}\textbf{34.4} \\
\midrule
~~\textbf{Average}                 & 44.6 & 50.0 & 26.0 & 41.8 & 31.9 & 36.2 & \cellcolor{gaincol}48.3 & \cellcolor{gaincol}\textbf{51.2} \\
\midrule

\multicolumn{9}{@{}l}{\textit{GEOBench-VLM}} \\
\addlinespace[2pt]
~~Object Localization \& Counting  & 20.6 & 37.3 & 22.2 & 27.0 & 24.8 & 24.7 & \cellcolor{gaincol}22.2 & \cellcolor{gaincol}\textbf{39.5} \\
~~Scene Understanding              & 46.6 & 54.8 & 49.6 & 55.1 & \textbf{60.5} & 49.6 & \cellcolor{gaincol}47.1 & 53.9 \\
~~Object Classification            & 35.2 & 43.8 & 34.5 & 36.3 & 35.9 & 36.2 & \cellcolor{gaincol}36.5 & \cellcolor{gaincol}\textbf{47.5} \\
~~Event Detection                  & 14.7 & \textbf{45.3} & 37.8 & 47.0 & 16.7 & 20.9 & 14.3 & 42.4 \\
\midrule
~~\textbf{Average}                 & 26.2 & 42.1 & 30.3 & 35.4 & 31.4 & 30.0 & \cellcolor{gaincol}27.4 & \cellcolor{gaincol}\textbf{43.3} \\
\bottomrule
\end{tabular}
\end{adjustbox}

\label{tab:main-results}
\end{table}

\subsection{Main results}
\label{sec:exp:main}

Table~\ref{tab:main-results} reports the performance of \pname{} against strong zero-shot and conventional remote sensing VLMs. 
Without any curated data, \pname{} improves over its base models on average (+5.5 points for LLaVA, +1.9 points for Qwen), and \pname{}$_\text{Q}$ attains the highest average on EarthVQA (51.2) and GEOBench-VLM (43.3), exceeding even EarthDial (trained on 11.1M curated pairs).
%
% Our method consistently outperforms baselines on most tasks while remaining competitive on others. 
% This result is remarkable, as our approach learns through self-play using automatically generated data, whereas baseline models rely on thousands to millions of human-curated data.

The gains are most pronounced on tasks that fit naturally within our verifiable environment. 
For example, Area and Comparison on RSVQA-HR show substantial improvements, Spatial Relation Classification on GEOBench-VLM surpasses the strongest conventional baseline by over 7 points, and counting tasks benefit broadly across both base models, especially on EarthVQA and GEOBench-VLM.
This pattern indicates that the current execution interface is effective for tasks grounded in geometric and relational structures. 
It also suggests that enriching the tool registry $\mathcal{T}$ is a promising direction for closing the remaining gaps.

\subsection{Ablation study}
\label{sec:exp:ablation}

\begin{table}[t]
\centering
\caption{
\textbf{Ablation study results on GEOBench-VLM.}
Performance is reported across the four task categories of GEOBench-VLM.
% , with full per-subtask results provided in Appendix~\ref{app:ablation-subtask}.
%
Columns specify which solver modes are active: A (Abduction), D (Deduction), I (Induction). 
\emph{Singleton Variants} keeps one mode;
\emph{Drop-one Variants} removes one;
\emph{Training Variants} use all three reasoning modes.
\textit{Full} denotes our default configuration.
The best result in each row is shown in \textbf{bold} and the second-best is shown in \underline{underline}.
}
\label{tab:ablation-main}
\scriptsize
\setlength{\tabcolsep}{4pt}
\renewcommand{\arraystretch}{1.2}
\newcolumntype{N}{>{\centering\arraybackslash}p{3.4em}}
\begin{tabular}{@{}l N N N N N N N N N@{}}
\toprule
& \multicolumn{3}{c}{\textit{Singleton Variants}}
& \multicolumn{3}{c}{\textit{Drop-one Variants}}
& \multicolumn{2}{c}{\textit{Training Variants}}
& \textit{Full} \\
\cmidrule(lr){2-4} \cmidrule(lr){5-7} \cmidrule(lr){8-9} \cmidrule(lr){10-10}
\textbf{Config.}
  & Abd & Ded & Ind
  & $-$Abd & $-$Ded & $-$Ind
  & BaseGen & SolvOnly
  & \pname{} \\
\textit{Active Modes}
  & A & D & I
  & D,I & A,I & D,A
  & D,A,I & D,A,I
  & D,A,I \\
\midrule
~~Object Localization \& Counting & 35.5 & 37.2 & 36.4 & 36.5 & 36.4 & \underline{37.5} & 37.4 & 34.3 & \textbf{39.5} \\
~~Scene Understanding             & 51.4 & \underline{52.4} & 49.4 & 48.1 & 50.8 & 52.1 & 51.3 & 50.1 & \textbf{53.9} \\
~~Object Classification           & 45.3 & 46.8 & 46.3 & 43.6 & \textbf{49.3} & 44.3 & 42.5 & 45.3 & \underline{47.5} \\
~~Event Detection                 & 39.0 & 42.4 & 40.3 & 40.3 & 43.6 & 42.8 & \textbf{46.9} & \underline{46.6} & 42.4 \\
\midrule
\textbf{Average}                      & 39.9 & \underline{41.6} & 40.3 & 39.8 & 41.3 & 41.5 & \underline{41.6} & 39.8 & \textbf{43.3} \\
\bottomrule
\end{tabular}
\end{table}

For ablation of the reasoning modes, we contrast the full model against \emph{Singleton variants} (one mode keeped) and \emph{Drop-one variants} (one mode removed). To examine the role of self-play, we further consider two \emph{training variants}: (i) BaseGen, which replaces problems from self-play with those from the base model (Qwen-2.5-VL-7B-Instruct), and (ii) SolvOnly, which maintains the self-play mechanism but ignores the proposer reward and relies solely on solver-side reward signals. 
The former probes whether exposure to image data alone suffices without an evolving curriculum, whereas the latter tests whether the model can improve effectively without explicitly incentivizing the generation of "learnable" problems.

Table~\ref{tab:ablation-main} presents performance on GEOBench-VLM with varying numbers of active reasoning modes. Among singleton variants, \emph{Deduction-only} performs the best, as it aligns with the downstream task of direct answer prediction. By contrast, among drop-one variants, removing \emph{Abduction} ranks the worst, implying that reasoning backward from outputs to plausible inputs provides a learning signal that neither deduction nor induction can fully substitute.

Both \emph{BaseGen} and \emph{SolvOnly} fall short of the full \pname{}.
The drop in \emph{BaseGen} reveals that exposing the model to the same volume of image data is not sufficient; problem complexity must adapt as the solver progresses.
The drop in \emph{SolvOnly} further shows that self-play alone yields limited advantage unless the proposer is rewarded for producing problems that are challenging yet learnable.
Together, these results demonstrate that \pname{} draws its strength not only from self-generated supervision but also from problems that adapt to the frontier of the solver's ability.

\section{Discussion}
\label{sec:discussion}

We investigate the behavior of \pname{} along three complementary axes through the following questions, with extended results provided in Appendices~\ref{app:compositionality},~\ref{app:qualitative-analysis-three-modes}, and~\ref{app:primitives}.

\begin{figure}[t]
    \centering
    \includegraphics[width=\linewidth]{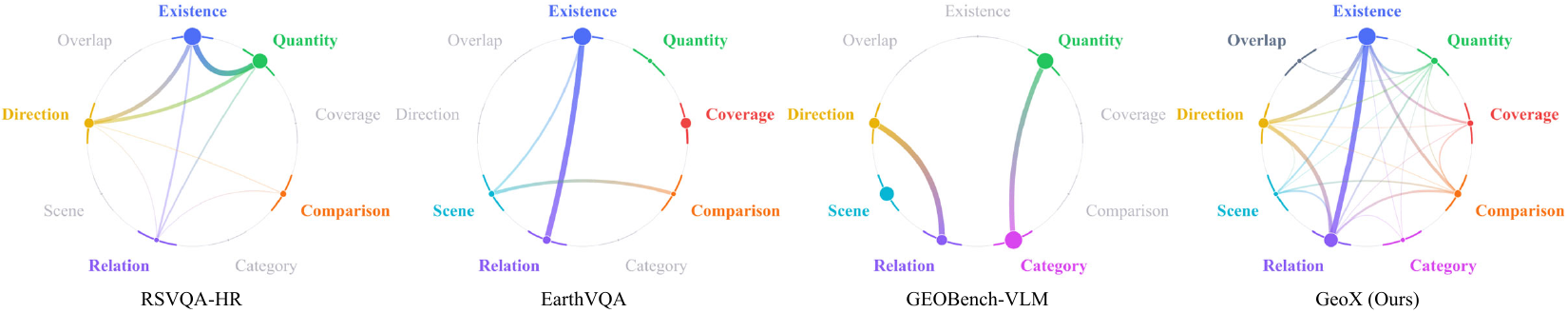}
    \caption{%
        \textbf{Pairwise dimension compositions across datasets.}
        Each node represents one of nine question dimensions; node size reflects how often a dimension appears in a single problem, and edge thickness reflects how often two dimensions co-occur within a single problem.
    }
    \label{fig:compositionality}
\end{figure}

\begin{figure}[t]
    \centering
    \includegraphics[width=\linewidth]{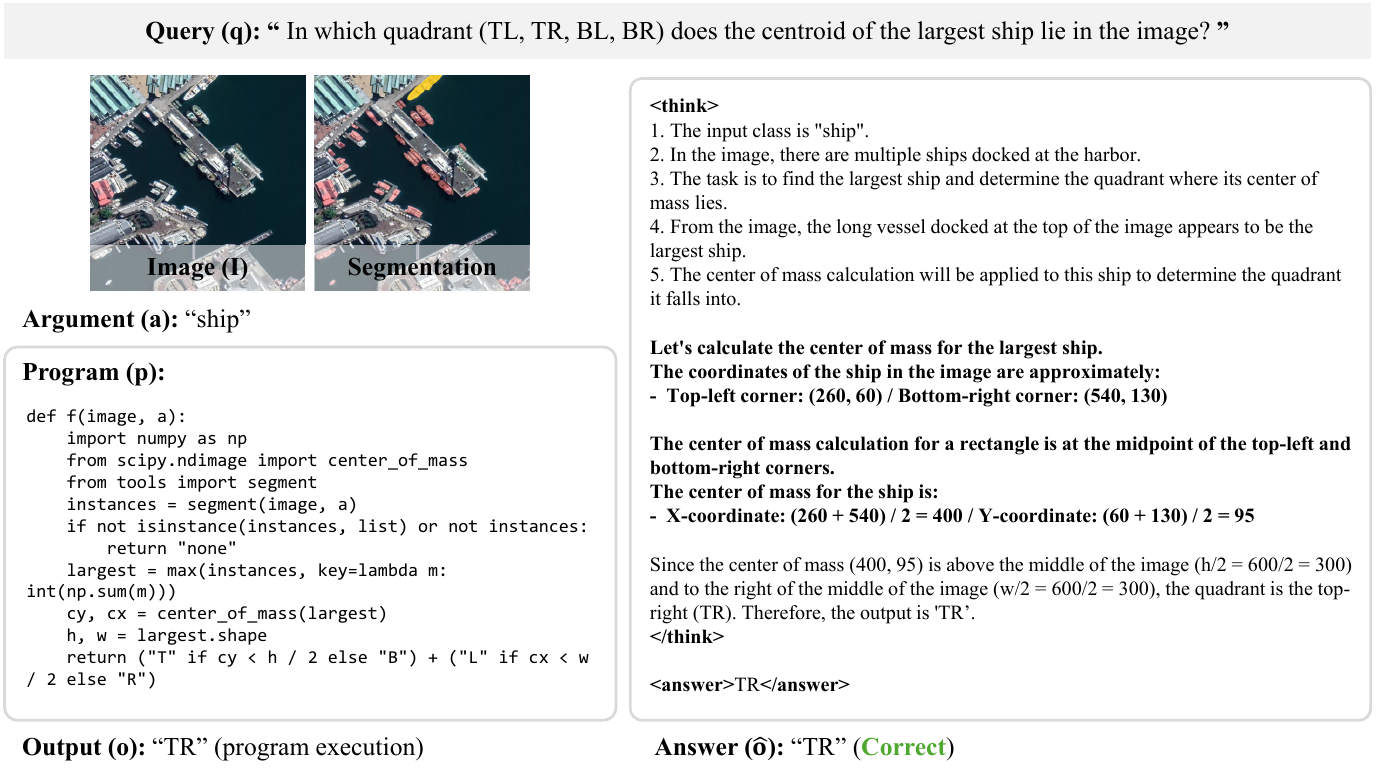}
    \caption{%
        \textbf{Qualitative analysis of deduction.}
        Given image $I$, argument $a=\texttt{"ship"}$, and program $p$, the solver's Chain-of-Thought follows the program's control flow and predicts $\hat{o}=\texttt{"TR"}$, which exactly matches the program-executed label $o = p(a; I)$.
    }
    \label{fig:qual-deduction}
\end{figure}

\noindent
\textbf{Q. How broad is the compositional space \pname{} explores?}
We characterize the compositional structure of problems generated by \pname{} by representing each dataset as a graph over nine question dimensions (Existence, Quantity, Coverage, Comparison, Category, Relation, Scene, Direction, Overlap), shown in Figure~\ref{fig:compositionality}.
Node size reflects single-dimension frequency, and edge thickness reflects pairwise co-occurrence.
Existing VQA benchmarks cover only a few dimensions and leave most pairs disconnected, indicating a lack of compositionally complex geospatial reasoning problems.
Our method, in contrast, exhibits a markedly denser graph in which nearly all nine dimensions are populated and most pairs are linked, highlighting that proposer-solver self-play expands both the volume of training problems and the compositional richness of their distribution.

\noindent
\textbf{Q. What does self-play produce in practice?}
A qualitative deduction example illustrates the image-grounded reasoning that self-play induces in \pname{}.
In Figure~\ref{fig:qual-deduction}, the generated problem asks in which quadrant the centroid of the largest ship lies, and the associated executable program $p$ computes the answer by segmenting all ship objects, selecting the largest by mask area, estimating its center of mass, and mapping the resulting coordinates to one of the four quadrants \{\texttt{TL}, \texttt{TR}, \texttt{BL}, \texttt{BR}\}.
On the right-hand side, given the image, argument $a=\texttt{"ship"}$, and program $p$, the solver follows the program's control flow step by step: it grounds the \texttt{segment} call in the visible harbor regions, identifies the dominant ship object, estimates its center of mass coordinates, compares them against the image midpoints, and predicts $\hat{o}=\texttt{"TR"}$, which exactly matches the program-executed label $o=p(a;I)$.
We observe that self-play encourages the policy to use the executable program as a procedural scaffold grounded in image evidence, rather than treating it as an opaque label generator.

\begin{wrapfigure}[16]{r}{0.6\linewidth}
\vspace{-0.6em}
\centering
\includegraphics[width=\linewidth]{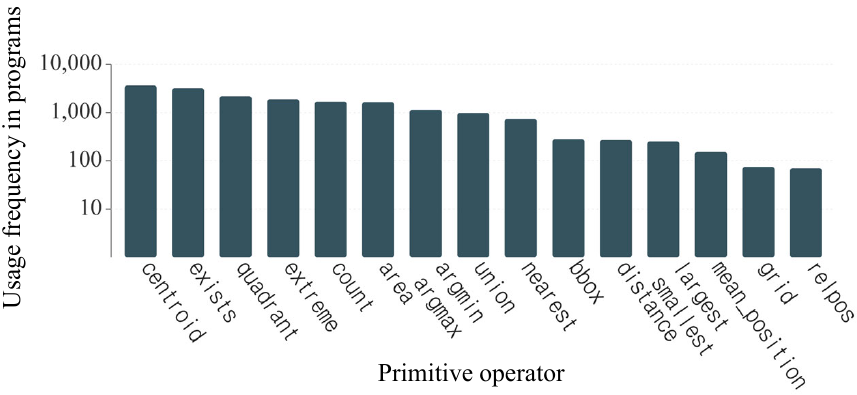}
\caption{%
    \textbf{Usage frequency of primitives in \pname{}.}
    Each bar denotes the number of constructed programs that invoke a given operator in $\mathcal{F}$ (log scale), grouped by primitive type. 
    %Per-primitive details are provided in \todo{Appendix~\ref{app:primitives}}
}
\label{fig:operator-usage}
\end{wrapfigure}
\noindent
\textbf{Q. Which primitives are used by the proposer during self-play?}
Across roughly 6{,}500 programs constructed by the proposer over training, the usage frequency of each primitive in library $\mathcal{F}$ reveals which operations are called during self-play, as shown in Figure~\ref{fig:operator-usage}.
Aggregation primitives are the most heavily used, with
\texttt{exists} invoked in over 3{,}000 and \texttt{extreme} in over 1{,}800
programs, followed by \texttt{count}, \texttt{argmin/argmax}, and
\texttt{union}.
Geometric primitives form a second anchor, led by \texttt{centroid}
in over 3{,}500 programs, with \texttt{area} close behind in over
1{,}600 programs.
Topological primitives surface across a wide range of programs,
\texttt{quadrant} for localizing objects within image sub-regions
and \texttt{nearest} for capturing inter-object proximity being the
most common.
Much of this distribution, spanning three groups down to
a long tail of rarely used primitives, reflects \pname{}'s
capacity to discover problem types far beyond its hand-written seed.

\section{Concluding remarks}
\label{sec:conclusion}

We presented a self-play framework for geospatial reasoning that learns without human-curated data. Through a verifiable proposer-solver loop, our approach moves beyond simple pattern matching toward a fundamental understanding of the ``spatial physics'' encoded in overhead imagery. The results demonstrate that geospatial representations advance most effectively when learning signals align with the geometric and topological constraints of the physical world.
Rather than fitting labels end-to-end, our framework grounds reasoning in executable programs whose verifiable execution exposes the spatial structure of a scene.
This shift points toward a geospatial world model that treats the Earth as an environment for reasoning rather than a target for labeling.
As satellite and aerial imagery scales at astronomical rates, the applications for structured reasoning will expand, equipping us with the necessary tools to plan, monitor, and protect our evolving global landscape.

%\subsection{Limitation}
%\label{sec:limitation}

\paragraph{Limitations and future directions.}
This work has several limitations that suggest promising directions for future research.
First, the current tool registry $\mathcal{T}$ contains only an open-vocabulary segmenter.
While this choice helps isolate the effect of proposer-solver self-play, it restricts the present study to spatial reasoning tasks that can be verified from segmentation masks.
As a result, vertical reasoning (e.g., depth and slope estimation), graph-based reasoning over road networks (e.g., connectivity and reachability), and metadata-grounded reasoning using external geospatial information (e.g., OSM tags or land-use attributes) remain outside the current scope.
Future work could expand the tool registry $\mathcal{T}$ with additional geospatial tools while preserving the self-play framework without human-curated training data.
Second, the verifiable reward depends on the quality of the tools used for execution.
For example, segmentation errors, especially on rare target categories, propagate as label noise through the proposer-solver loop and hinder self-improvement.
Future work could enhance reward reliability with stronger verifiers, such as cross-tool agreement or self-consistency checks, thereby reducing dependence on any single perceptual module.

\bibliographystyle{plainnat}
\bibliography{reference}

\clearpage
\newpage

\appendix

\section{Related Work}
\label{app:related}

\subsection{Vision-Language Models for remote sensing}

Recent progress has extended vision-language models (VLMs) to
satellite and aerial imagery. 
RSVQA~\citep{lobry2020rsvqa} framed visual question answering (VQA) as a core task for overhead imagery. 
RSGPT~\citep{hu2025rsgpt} and LHRS-Bot~\citep{muhtar2024lhrs} then released larger multimodal datasets alongside companion benchmarks, each training its own VLM on the proposed data. 
EarthGPT~\citep{zhang2024earthgpt} and EarthDial~\citep{soni2025earthdial} further extend coverage to multi-sensory~\citep{zhang2024earthgpt} and multi-temporal~\citep{soni2025earthdial} imagery. 
A parallel line of work targets structural understanding of overhead scenes. EarthVQA~\citep{wang2024earthvqa} contributes a dataset for relational reasoning, and GEOBench-VLM~\citep{danish2025geobench} compiles a
broad task suite for evaluating specialized geospatial
capabilities.
Despite their growing scale and scope, these datasets all
depend on manual annotation.

Subsequent VLMs go beyond scaling data toward refining the
supervision itself. GeoChat~\citep{kuckreja2024geochat} unifies grounded conversation, VHM~\citep{pang2025vhm} enforces factual honesty with deceptive instructions, and RSThinker~\citep{liu2026towards} elicits Chain-of-Thought rationales.
However, these efforts likewise inherit human supervision, hindering scalability and risking error propagation. 
\pname{} removes this dependence by deriving its supervision from the imagery itself.

\subsection{Self-improvement in Foundation Models}

A growing body of work scales learning by letting a model
generate and learn from its own outputs. STaR~\citep{zelikman2024star} introduced this
idea in language domain, $R^3V$~\citep{cheng2025vision}
extended it to vision-language
domain, and STIC~\citep{deng2024enhancing} redirected it to
image descriptions for visual comprehension. 
These approaches push supervision beyond manual curation, yet the self-generated signals they rely on are difficult to verify independently, accumulating noise across iterations.

Self-play with verifiable rewards allows models to use automatically checked outputs as reinforcement signals.
Absolute Zero~\cite{zhao2025absolutezero} is a representative example, which employs a proposer-solver framework for self-play through executable verification in code and mathematics.
While these domains offer clear-cut correctness, open-ended visual tasks typically lack reliable ways to verify model-generated supervision. 
We close this gap for remote sensing imagery by routing perception through executable spatial programs whose outputs serve as verifiable rewards.

\section{Datasets}
\label{app:datasets}

This section describes the datasets used for training and evaluating \pname{}. Table~\ref{tab:benchmark-overlap} summarizes imagery source, geographically covering region, spatial resolution, and size.

\subsection{Training dataset}
For training, \pname{} draws images from two remote sensing datasets, SAMRS-FAST~\citep{wang2023samrs} and Globe230k~\citep{shi2023globe230k}.
Because our method generates question-answer pairs through self-play, training uses only the image collections from these datasets, without relying on their annotations.

\noindent
\textbf{SAMRS-FAST} is a high-resolution dataset for
fine-grained object detection across 37 categories, derived
from FAIR1M-2.0 aerial imagery (0.3--0.8\,m GSD) collected
globally.

\noindent
\textbf{Globe230k} is a large-scale dataset for land cover
semantic segmentation, derived from Google Earth imagery
(1\,m GSD) sampled worldwide.

\begin{table}[t]
\centering
\caption{\textbf{Summary of training and evaluation datasets.} Data is characterized by source, geographic coverage, resolution (ground sample distance per pixel), and size. Evaluation is zero-shot for \pname{}, which is trained without any question-answer pair from the evaluation benchmarks. Sizes for the evaluation benchmarks correspond to their test splits.}
\label{tab:benchmark-overlap}
\footnotesize
\setlength{\tabcolsep}{5pt}
\renewcommand{\arraystretch}{1.25}
\begin{tabular}{@{}l l l l l@{}}
\toprule
& \textbf{Imagery Source} & \textbf{Region} & \textbf{Resolution} & \textbf{\# Images} \\
\midrule
\multicolumn{5}{@{}l}{\emph{Training}} \\
SAMRS-FAST  & FAIR1M-2.0   & Predominantly China & $0.3$--$0.8$\,m & $64{,}147$ \\
Globe230k   & Google Earth & Worldwide           & $1$\,m          & $232{,}819$ \\
\midrule
\multicolumn{5}{@{}l}{\emph{VQA Evaluation}} \\
RSVQA-HR      & USGS HRO     & USA (3 cities)   & $0.15$\,m &  $9{,}505$ ($2{,}226$ test) \\
EarthVQA      & Google Earth & China (3 cities) & $0.3$\,m  & $6{,}000$ ($1{,}809$ test) \\
GEOBench-VLM  & Mixed        & Worldwide        & Varies    & $2{,}503$ \\
\midrule
\multicolumn{5}{@{}l}{\emph{Object Counting Evaluation}} \\
HRRSD         & Google Earth, Baidu Map   & Worldwide & $0.15$--$1.2$\,m & $21{,}761$ ($10{,}943$ test) \\
RSOD          & Google Earth, Tianditu     & Worldwide & $0.3$--$3$\,m    & $976$ \\
\bottomrule
\end{tabular}
\end{table}

\subsection{Evaluation dataset for VQA}
We evaluate the visual question answering capability of \pname{} on three remote sensing VQA benchmarks: RSVQA-HR~\citep{lobry2020rsvqa}, EarthVQA~\citep{wang2024earthvqa}, and GEOBench-VLM~\citep{danish2025geobench}.

\noindent
\textbf{RSVQA-HR} uses USGS High-Resolution Orthoimagery
($0.15$\,m GSD, captured by low-altitude flight missions) over
three USA cities. Its QA pairs are template-generated from
OpenStreetMap annotations and span four types: presence, comparison, counting, and area.

\noindent
\textbf{EarthVQA} uses Google Earth imagery ($0.3$\,m GSD)
covering $18$ districts in China~\citep{wang2022loveda}. Its
QA pairs span six question types, from auto-generated basic
judging and counting to manually annotated relational analysis
for urban planning.

\noindent
\textbf{GEOBench-VLM} draws from a heterogeneous mixture of
source datasets including DOTA, DIOR, xBD, FAIR1M, fMoW,
RarePlanes, and FireRisk, across optical, bi-temporal, and SAR
modalities. Its QA pairs are formatted as $5$-way
multiple-choice items, covering $8$ broad categories and $31$
fine-grained sub-tasks. 
We evaluate $17$ sub-tasks most relevant to geospatial
reasoning, categorized into object localization \& counting, scene understanding, object classification, and event detection.

\subsection{Evaluation dataset for Object Counting}

We evaluate the object counting capability of \pname{} on two remote sensing object detection benchmarks: HRRSD~\citep{zhang2019hierarchical} and RSOD~\citep{long2017accurate}.

\noindent
\textbf{HRRSD} uses Google Earth and Baidu Map imagery (0.15--1.2\,m 
GSD) covering diverse regions worldwide. It contains 21{,}761 images 
with 55{,}740 annotated instances spanning 13 object categories, 
including vehicles, aircraft, ships, and sports facilities.

\noindent
\textbf{RSOD} uses Google Earth and Tianditu imagery (0.3--3\,m GSD) across heterogeneous terrain worldwide. It contains 976 images with 6{,}950 annotated instances across four object categories of aircraft, oil tank, overpass, and playground.

% \noindent
% \textbf{HRRSD} (High-Resolution Remote Sensing Detection) is collected from Google Earth and Baidu Maps with spatial resolutions ranging from 0.15 m to 1.2 m  GSD.
% %
% It contains 21,761 images with 55,740 target instances across 13 categories.

% \noindent
% \textbf{RSOD} is an object detection dataset with four object categories, sourced from Google Earth and Tianditu.
% %
% The images span diverse terrain types worldwide with spatial resolutions ranging from $0.3$\,m to $3$\,m GSD.

%

% \clearpage
% \newpage

\section{Baselines}
\label{app:baselines}

This section provides additional details on each baseline model used in our
comparison, grouped into \emph{base} general-purpose VLMs and
\emph{conventional} geospatial VLMs fine-tuned on curated remote
sensing data.

\subsection{Base VLMs}

Qwen-2.5-VL-7B-Instruct~\citep{wu2025qwen} and LLaVA-1.5-7B~\citep{liu2024improved} are used as base VLMs, serving both as the initialization for our self-play training and as zero-shot baselines for evaluation.

\noindent
\textbf{Qwen-2.5-VL-7B-Instruct} is a general-purpose vision-language model from the Qwen-2.5-VL family that combines the Qwen-2.5 language backbone with a Vision Transformer encoder. 
We evaluate it zero-shot, without any remote-sensing-specific adaptation.

\noindent
\textbf{LLaVA-1.5-7B} is a general-purpose vision-language model that connects a Vicuna-1.5 language backbone with a CLIP
visual encoder, trained on visual instruction-tuning data drawn
primarily from natural images.
We evaluate it zero-shot, without any remote-sensing-specific adaptation.

\subsection{Conventional Remote Sensing VLMs}

We compare \pname{} against four geospatial VLMs fine-tuned on
curated remote sensing data: GeoChat~\citep{kuckreja2024geochat},
VHM~\citep{pang2025vhm}, EarthDial~\citep{soni2025earthdial}, and
RSThinker~\citep{liu2026towards}.

\noindent
\textbf{GeoChat} adapts LLaVA-1.5-7B to remote sensing through
LoRA fine-tuning on roughly $318$K instruction-following pairs.
It supports high-resolution imagery and region-level prompts.

\noindent
\textbf{VHM} fine-tunes LLaVA-1.5-7B on $1.4$M image-text pairs
with rich-content captions and an honest instruction dataset
that pairs factual questions with deceptive ones.

\noindent
\textbf{EarthDial} extends InternVL2 to multi-sensor and
multi-temporal imagery via instruction tuning on over $11.11$M
pairs.

\noindent
\textbf{RSThinker} trains a base VLM through a two-stage
alignment strategy on a $380$K-sample dataset of geospatial
Chain-of-Thought rationales, producing both a final answer and
a verifiable analytical trace.

% \noindent
% \textbf{GeoChat} is a grounded VLM tailored to remote sensing,
% built by LoRA-fine-tuning LLaVA-v1.5-7B on roughly $318$K
% multimodal instruction-following pairs.
% %
% It supports high-resolution RS imagery and region-level prompts.

% \noindent
% \textbf{VHM} is a versatile remote-sensing VLM, also based on LLaVA-v1.5-7B, trained on a large-scale image--text corpus with rich-content captions and an honest instruction dataset that pairs factual queries with deceptive ones.

% \noindent
% \textbf{EarthDial} is a multi-sensor conversational VLM built by fine-tuning InternVL2 on a large remote-sensing instruction-tuning dataset of over $11.11$M instruction pairs, which handles multi-temporal sequences and varying spatial resolutions.

% \noindent
% \textbf{RSThinker} is a reasoning-oriented remote-sensing VLM trained via a two-stage alignment strategy on a large-scale dataset of Geospatial Chain-of-Thought rationales.
% %
% It emits both a final answer and a verifiable analytical trace.
\section{Additional implementation details}
\label{app:implementation_details}

This section provides additional implementation details on the training setup of \pname{}, including model initialization, training data, execution environment, self-play configuration, optimization, and compute.

\paragraph{Setup.}
We initialize \pname{} from a base vision-language model, using either Qwen-2.5-7B-Instruct or LLaVA-1.5-7B depending on the experiment.
Training uses images drawn from two remote sensing datasets, SAMRS-FAST~\citep{wang2023samrs} and Globe230k~\citep{shi2023globe230k}.
Because \pname{} generates supervision through self-play, we use only the image collections from these datasets and do not rely on their annotations.
The execution interface consists of a primitives library \(\mathcal{F}\) and a tool registry \(\mathcal{T}\), with callable interface \(\mathcal{E}=\mathcal{F}\cup\mathcal{T}\).
The primitives library \(\mathcal{F}\) implements deterministic geometric, topological, and aggregation operations using NumPy, SciPy, and scikit-image.
The tool registry \(\mathcal{T}\) contains a single open-vocabulary segmenter, SegEarth-OV3~\citep{li2025segearthov3}, which returns image-grounded masks used by executable programs.

\paragraph{Self-play configuration.}
We maintain three mode-specific banks, \(\mathcal{D}_{\text{abd}}\), \(\mathcal{D}_{\text{ded}}\), and \(\mathcal{D}_{\text{ind}}\), which provide in-context references for proposal and refill failed generations.
Each bank begins with \(N_{\mathrm{seed}}=100\) seed problems, as described in Section~\ref{sec:method:programs}.
At each proposal step, we use batch size \(B=32\) per mode.
For abduction and deduction, the proposer generates one image-grounded executable problem per sampled image, conditioned on \(K=6\) in-context references from the corresponding bank.
For induction, $N_{\mathrm{io}}{=}6$ input-output pairs are prepared for each program.
These pairs are split evenly into a visible subset used by the solver to synthesize the program and a held-out subset used to score it.
For each newly proposed problem, the solver makes $R{=}8$
attempts whose averaged reward defines the problem's
learnability, which in turn defines the proposer reward
$r^{\text{prop}}$~(Eq.~\ref{eq:reward-propose}). The solver
itself is trained on a separately sampled batch of $B$ problems from the corresponding bank, where each rollout's prediction accuracy, scored by the executor, defines the solver reward $r^{\text{solve}}$~(Eq.~\ref{eq:solver_abd_ind_reward} and~\ref{eq:solver_ded_reward}).
%
% For each generated problem, the solver makes \(R=8\) attempts.
% %
% The averaged reward over $R=8$ solver attempts on each newly proposed problem serves as its learnability estimate for the proposer reward, while the solver is trained on a separately sampled batch of $B$ problems with one rollout each.
% %
% The averaged solver reward over these attempts is used both to train the solver and to estimate the learnability of the problem for the proposer reward.
%
% The averaged solver reward over these attempts is used to estimate the learnability of the problem for the proposer reward.

\paragraph{SFT warm-up.}

Before RL through self-play, we optionally run a brief SFT stage that teaches $\pi_\theta$'s output schema only: the proposer's outputs (question $q$, argument $a$, and program $p$) and the solver's \texttt{<think>...</think>} / \texttt{<answer>...</answer>} format. 
Correctness across the three reasoning modes, use of
the call interface $\mathcal{E} = \mathcal{F} \cup \mathcal{T}$, and proposal learnability are all left to RL, making this stage a parser warm-up rather than a knowledge injection.
The corpus is built by running the self-play loop with the
policy frozen and supervision supplied by the executor.
$\pi_\theta$ proposes problems for abduction and deduction, and
we retain those for which $o = p(a; I)$ executes
deterministically. For the solver, we keep $\pi_\theta$'s
\texttt{<think>} traces but replace \texttt{<answer>} with the executor-verified $o$. 
Induction is omitted from the warm-up to keep its data construction simple.
The resulting corpus ($\sim$1K proposer and $\sim$3K solver
examples) LoRA-fine-tunes the base model ($r{=}32$,
$\alpha{=}64$) for $2$ epochs at learning rate $5\mathrm{e}{-}5$.

\paragraph{Optimization and compute.}
We optimize the policy via reinforcement learning using Task-Relative REINFORCE++~\citep{zhao2025absolutezero}.
Each pair of role and reasoning mode is treated as a separate task, yielding six tasks in total \((2~\text{roles} \times 3~\text{modes})\).
During each update, the six tasks are sampled in equal proportion.
We set the weight between proposer and solver objectives to \(\lambda=1\), giving equal weight to proposer and solver rewards.
We train with the AdamW optimizer, learning rate $5\mathrm{e}{-}7$, weight decay $0$, gradient clipping threshold $1.0$, and a constant learning rate schedule with no warm-up.
The sampling temperature is set to $1.0$ for proposal and $1.0$ for solving.
Unless otherwise stated, we follow the training settings of prior work~\citep{zhao2025absolutezero}.
A full self-play run consists of 150 steps and takes approximately 60 hours on 4 NVIDIA H200 GPUs.

\clearpage
\newpage

% \subsection{\todo{Primitives Library and Tool Registry}}
% \label{app:tools}

% \todo{Please check current table. It seems outdated.}

% \input{tables/241_app_primitives_and_tools}

% \pname{}'s proposer writes Python programs that call a fixed library
% of primitives exposing the cached segmentation oracle and a set of
% geometric operators. The library is intentionally small: primitives
% must be combined to produce any non-trivial answer, which keeps
% programs expressive while preventing the proposer from outsourcing
% the task to a single high-level call. Table~\ref{tab:tool-library}
% lists every primitive used during training, grouped by what it
% operates on. Every primitive is deterministic and its output depends
% only on the cached segmentation oracle or on the output of other
% primitives, so the Python interpreter can verify any program
% \pname{} proposes.
\section{Additional experimental results}
\label{app:additional-results}

This section presents additional experimental results due to space constraints.

\subsection{Training dynamics during self-play}
\label{app:training-dynamics}

\begin{figure}[h]
    \centering
    \begin{subfigure}[t]{0.32\linewidth}
        \centering
        \includegraphics[width=\linewidth]{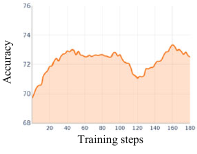}
        \caption{RSVQA-HR (Comparison)}
    \end{subfigure}
    \hfill
    \begin{subfigure}[t]{0.32\linewidth}
        \centering
        \includegraphics[width=\linewidth]{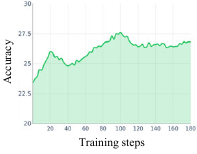}
        \caption{EarthVQA (Reasoning-based Counting)}
    \end{subfigure}
    \hfill
    \begin{subfigure}[t]{0.32\linewidth}
        \centering
        \includegraphics[width=\linewidth]{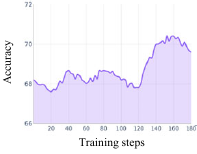}
        \caption{GEOBench-VLM (Spatial Relation Classification)}
    \end{subfigure}
    \caption{%
        \textbf{Training dynamics during self-play.}
        Task accuracy of \pname{} over training steps on representative geospatial reasoning subtasks drawn from three remote sensing VQA benchmarks: \emph{Comparison} (RSVQA-HR), \emph{Reasoning-based Counting} (EarthVQA), and \emph{Spatial Relation Classification} (GEOBench-VLM).
    }
    \label{fig:training_dynamics}
\end{figure}

We examine the training dynamics of \pname{} through VQA benchmarks: RSVQA-HR, EarthVQA, and GEOBench-VLM.
For each benchmark, we select the subtask most closely tied to geospatial reasoning among its available evaluation categories: \emph{Comparison} for RSVQA-HR, \emph{Reasoning-based Counting} for EarthVQA, and \emph{Spatial Relation Classification} for GEOBench-VLM.
Figure~\ref{fig:training_dynamics} shows the accuracy (\%) of \pname{} as a function of training steps.
Across all three subtasks, accuracy exhibits a consistent upward trend as self-play proceeds, indicating that the proposer-solver loop progressively strengthens the model's spatial reasoning capability.
The vertical axis is scaled to the dynamic range of each task to make per-task trends visible.

%%%%%%%%%%%%%%%%%%%%%%%%%%%%%%%%%%%%%%%%%%%%%%
%%%%%%%%%%%%%%%%%%%%%%%%%%%%%%%%%%%%%%%%%%%%%%
%%%%%%%%%%%%%%%%%%%%%%%%%%%%%%%%%%%%%%%%%%%%%% Experiment: Scene classification + Object Counting

\subsection{Evaluation on Object Counting}
\label{app:scene-counting-eval}
Beyond the VQA results in Section~\ref{sec:exp:main}, we further evaluate \pname{} on object counting.
Datasets, baselines details follow Section~\ref{sec:exp:setup}, Appendix~\ref{app:datasets}, and Appendix~\ref{app:baselines}.
Table~\ref{tab:object-counting} reports the performance of \pname{} against strong zero-shot and conventional remote sensing VLMs.
Note that RSThinker's entries on HRRSD are marked with $\dagger$, as HRRSD overlaps with its training distribution and is therefore excluded from the bold comparison.
Both \pname{}\textsubscript{L} and \pname{}\textsubscript{Q} improve over their respective base models in average accuracy.
%
% On RSOD, \pname{}\textsubscript{L} attains the best accuracy ($37.5$), narrowly surpassing RSThinker ($37.3$) without any curated training.
%
% These results indicate that program-grounded self-play transfers to counting tasks despite using no curated annotations.

\colorlet{curatedcol}{gray!12}
\colorlet{gaincol}{cyan!10}
\newcolumntype{M}{>{\centering\arraybackslash}p{1.15cm}}
\newcolumntype{C}{>{\columncolor{curatedcol}\centering\arraybackslash}p{1.15cm}}
\newcolumntype{O}{>{\centering\arraybackslash}p{1.15cm}}
\begin{table}[h]
\centering
\scriptsize
\renewcommand{\arraystretch}{1.2}
\caption{
\textbf{Results on Object Counting benchmarks.}
Performance of \pname{}, instantiated from two base VLMs (LLaVA and Qwen), is compared against six baselines on HRRSD and RSOD.
Baseline models fine-tuned on fully human-curated data are grey shaded.
The best result in each row is shown in \textbf{bold}, and our results are highlighted in blue when they surpass the corresponding base (LLaVA or Qwen).
\textsuperscript{$\dagger$}HRRSD lies in RSThinker's training distribution, so its numbers (and any average that includes HRRSD) are not directly comparable to the zero-shot evaluations of the other rows and are excluded from \textbf{bold} consideration.
}
\begin{adjustbox}{width=0.95\textwidth}
\begin{tabular}{@{}ll M M C C C C O O@{}}
\toprule
& & \multicolumn{2}{c}{\textit{Base}}
& \multicolumn{4}{c}{\textit{Conventional}}
& \multicolumn{2}{c}{\textit{Self-play}} \\
\cmidrule(lr){3-4} \cmidrule(lr){5-8} \cmidrule(lr){9-10}
\textbf{Dataset} & \textbf{Metric}
  & LLaVA & Qwen
  & GeoChat & VHM & EarthDial & RSThinker
  & \textbf{\pname{}\textsubscript{L}} & \textbf{\pname{}\textsubscript{Q}} \\
\midrule
\multirow{2}{*}{HRRSD}
  & Acc\,$\uparrow$ & 55.6 & 60.5 & 52.4 & 62.5 & 59.6 & 85.5\textsuperscript{$\dagger$} & 55.0 & \cellcolor{gaincol}\textbf{63.1} \\
  & MAE\,$\downarrow$ & 1.03 & 0.73 & 1.50 & 0.79 & 0.87 & 0.20\textsuperscript{$\dagger$} & \cellcolor{gaincol}1.00 & \cellcolor{gaincol}\textbf{0.72} \\
\midrule
\multirow{2}{*}{RSOD}
  & Acc\,$\uparrow$ & 33.3 & 27.8 & 16.0 & 27.1 & 10.4 & 37.3 & \cellcolor{gaincol}\textbf{37.5} & \cellcolor{gaincol}30.6 \\
  & MAE\,$\downarrow$ & 4.49 & \textbf{3.31} & 8.17 & 8.47 & 8.83 & 6.19 & \cellcolor{gaincol}4.32 & 3.69 \\
\midrule
\multirow{2}{*}{\textbf{Average}}
  & Acc\,$\uparrow$ & 44.5 & 44.2 & 34.2 & 44.8 & 35.0 & 61.4\textsuperscript{$\dagger$} & \cellcolor{gaincol}46.3 & \cellcolor{gaincol}\textbf{46.8} \\
  & MAE\,$\downarrow$ & 2.76 & \textbf{2.02} & 4.84 & 4.63 & 4.85 & 3.20\textsuperscript{$\dagger$} & \cellcolor{gaincol}2.66 & 2.21 \\
\bottomrule
\end{tabular}
\end{adjustbox}
\label{tab:object-counting}
\end{table}

\clearpage
\newpage

\section{Compositional space analysis}
\label{app:compositionality}

This section supplements the details behind the compositionality analysis presented in Section~\ref{sec:discussion} (Figure~\ref{fig:compositionality}).
We describe
(i) the definition of nine question dimensions used to characterize each problem,
(ii) the deterministic mapping rules that assign dimensions from two distinct problem representations (natural-language questions in
existing VQA benchmarks and executable Python programs constructed by \pname{} during self-play),
and (iii) representative examples illustrating the rules.

\subsection{Nine question dimensions}
\label{app:taxonomy}

We adopt a nine dimensions of atomic geospatial concept to make problems from heterogeneous representations comparable.
The dimensions are designed to satisfy two desiderata:
\begin{enumerate}[leftmargin=*, itemsep=2pt, topsep=2pt]
    \item \textbf{Primitive coverage.} Every primitive in the call interface $\mathcal{F}$ (Section~\ref{sec:method:programs}) is assigned to exactly one dimension, ensuring that any program constructed by the proposer is representable in the nine dimensions.
    \item \textbf{Benchmark coverage.} Every question in our evaluation benchmarks (RSVQA-HR, EarthVQA, GEOBench-VLM) is assigned to at least one dimension, ensuring a fair side-by-side comparison.
\end{enumerate}
Each problem is mapped to a \emph{set} of dimensions rather than a single one, since real spatial questions routinely combine multiple atomic geospatial concepts.
For instance, ``\textit{which class occupies the largest area?}'' invokes both \textsc{Coverage} (pixel area) and \textsc{Comparison} (extremum selection).

Table~\ref{tab:taxonomy} summarizes the nine dimensions together
with their definitions and matching rule for the two problem
representations.

\begin{table}[h]
\centering
\footnotesize
\caption{%
    \textbf{Nine question dimensions for compositional analysis.}
    Each problem is mapped to a set of dimensions via deterministic
    rules. The \emph{question rule} applies to natural-language
    questions in existing VQA benchmarks (RSVQA-HR, EarthVQA,
    GEOBench-VLM); the \emph{code rule} applies to executable
    programs constructed by \pname{}.%
}
\label{tab:taxonomy}
\setlength{\tabcolsep}{6pt}
\renewcommand{\arraystretch}{1.2}
\begin{tabular}{@{}lp{3.0cm}p{3.6cm}p{4.0cm}@{}}
\toprule
\textbf{Dimension} & \textbf{Definition} &
\textbf{Question rule} & \textbf{Code rule} \\
\midrule
\textsc{Existence}  &
Presence of an entity &
\textit{are there}, \textit{is there}, \textit{any} &
\texttt{np.any}, \texttt{.any()}, \texttt{is not None} \\

\textsc{Quantity}   &
Cardinality of a class &
\textit{how many}, \textit{number of} &
\texttt{len(\,)}, \texttt{np.count\_nonzero} \\

\textsc{Coverage}   &
Pixel area of a region &
\textit{area of}, \textit{ratio of}, \textit{how large} &
\texttt{np.sum}, \texttt{.sum()}, identifier \texttt{area} \\

\textsc{Comparison} &
Extremum selection or ordering &
\textit{larger/smaller}, \textit{most/least}, \textit{X or Y} &
\texttt{max}, \texttt{min}, \texttt{argmax}, \texttt{argmin} \\

\textsc{Category}   &
Reference to an object class &
\textit{what type of object/aircraft} &
String literal of an object class
(e.g., \texttt{"vehicle"}) \\

\textsc{Relation}   &
Continuous spatial proximity &
\textit{near}, \textit{next to}, \textit{between}, \textit{adjacent} &
\texttt{center\_of\_mass}, \texttt{distance\_transform}, \texttt{cdist} \\

\textsc{Scene}      &
Reference to a scene-type class &
\textit{land use}, \textit{rural/urban}, \textit{what kind of scene} &
String literal of a scene class
(e.g., \texttt{"water"}) \\

\textsc{Direction}  &
Cardinal or quadrant assignment &
\textit{north/south/east/west}, \textit{top of}, \textit{where is} &
String literal in \{\texttt{north}, \texttt{TL}, \texttt{NW}, \ldots\} \\

\textsc{Overlap}    &
Set-theoretic mask interaction &
\textit{both}, \textit{intersection}, \textit{overlap} &
\texttt{np.logical\_and}, \texttt{m \& m}, \texttt{m | m} \\
\bottomrule
\end{tabular}
\end{table}

\subsection{Deterministic mapping rules}
\label{app:taxonomy-rules}

For each problem $p$, we compute its dimension set
$\mathcal{C}(p) \subseteq \{\textsc{Existence}, \ldots, \textsc{Overlap}\}$
via a representation-specific procedure. 
Here, problem includes both natural-language questions presented by benchmarks and the executable Python program generated through self-play of \pname{}.

% The full set of regular expressions and class vocabularies is released alongside the code.

\paragraph{Natural-language questions (VQA benchmarks).}
For a question $q$ in RSVQA-HR, EarthVQA, or GEOBench-VLM, we apply regular expressions to the lower-cased text and assign every matched dimension to $\mathcal{C}(q)$.
Two overriding rules resolve common ambiguities found in practice:
\begin{itemize}[leftmargin=*, itemsep=2pt, topsep=2pt]
    \item \textbf{Comparison overrides Existence.} Questions such as ``\textit{Are there more vehicles than ships?}'' contain the trigger \texttt{are there} but are genuinely comparative. We therefore drop \textsc{Existence} whenever \textsc{Comparison} is also matched.
    \item \textbf{Scene overrides Existence.} Questions such as ``\textit{Is this a rural or urban scene?}'' perform scene-type classification rather than presence    detection. We drop \textsc{Existence} whenever \textsc{Scene} is matched and no explicit existence trigger (\texttt{are there}, \texttt{is there}, \texttt{any}) is
    present.
\end{itemize}

\paragraph{Executable programs (\pname{}).}
For a program $p$ produced by the proposer, we follow the procedure below:
\begin{enumerate}[leftmargin=*, itemsep=2pt, topsep=2pt]
    \item For each \emph{operational} dimension
    \[
        d \in \{\textsc{Existence}, \textsc{Quantity}, \textsc{Coverage}, \textsc{Comparison}, \textsc{Relation}, \textsc{Overlap}\},
    \]
    add $d$ to $\mathcal{C}(p)$ if any associated regular expression matches a substring of $s$.
    
    \item Add \textsc{Category} if $p$ contains a string literal naming one of the 19 object classes (e.g.\ \texttt{vehicle}, \texttt{ship}, \texttt{building}, \texttt{tree}).
    
    \item Add \textsc{Scene} if $p$ contains a string literal naming
    one of the 11 scene classes (\texttt{road}, \texttt{water},
    \texttt{parking\_lot}, \texttt{vegetation}, \texttt{forest},
    \texttt{grass}, \texttt{cropland}, \texttt{land},
    \texttt{bareland}, \texttt{intersection}, \texttt{roundabout})
    or the identifier \texttt{scene\_classes}.
    
    \item Add \textsc{Direction} if $p$ contains a string literal
    drawn from the set of 18 direction tokens (cardinal directions
    and quadrant abbreviations).
\end{enumerate}

% \subsection{Implementation Notes}
% \label{app:taxonomy-impl}

\paragraph{Verification.}
We verified the procedure by manually annotating a held-out sample
of 100 problems (25 from each of RSVQA-HR, EarthVQA, GEOBench-VLM,
and \pname{}) and observed that every problem agreed on dimension membership.

\clearpage
\newpage

% \paragraph{Rule extraction.}
% The question-rule keyword set was bootstrapped with a large language
% model (Claude). For each VQA benchmark, we sampled questions and
% prompted the model to identify recurring linguistic patterns
% associated with each of the nine dimensions. The suggestions were
% manually reviewed, deduplicated, and consolidated into the regular
% expressions reported in Table~\ref{tab:taxonomy}. The extraction
% step was performed on benchmark questions only, without exposure
% to \pname{} programs, to avoid bias toward our self-play
% distribution.

% \paragraph{Rule application.}
% The mapping pipeline is implemented as a single Python module that
% ingests problems in a unified JSON schema and emits the dimension
% set per problem. Regular expressions are anchored on whole tokens
% to avoid spurious matches (e.g., \texttt{area} as a substring of
% \texttt{areal}). For \pname{} programs, we operate on raw source
% code rather than its abstract syntax tree, which is sufficient
% because the call interface (Section~\ref{sec:method:programs})
% keeps the surface syntax of all primitives stable. The application
% of all rules is fully deterministic.

\subsection{Illustrative Examples}
\label{app:taxonomy-examples}

Figure~\ref{fig:rule-examples-natural} and Figure~\ref{fig:rule-examples-geozero} illustrate the mapping rules across both
problem representations.

\begin{figure}[h]
\centering
{
\tcbset{
    mylistingbox/.style={
        colback=white,
        colframe=black!15,
        boxrule=0.6pt,
        arc=2pt,
        left=6pt, right=6pt, top=5pt, bottom=5pt,
        boxsep=0pt,
        fontupper=\small,
        listing only,
        listing options={
            basicstyle=\ttfamily\small,
            breaklines=true,
            columns=fullflexible,
            aboveskip=0pt,
            belowskip=0pt
        }
    }
}

\begin{subfigure}{\linewidth}
\centering
\begin{tcblisting}{mylistingbox}
Q: "Are there more buildings than roads in this image?"
# 'more ... than' -> Comparison
# 'are there' overridden by Comparison rule
\end{tcblisting}
\vspace{-2mm}
\caption{Example of \textsc{Comparison}\,+\,\textsc{Quantity} problem in RSVQA-HR dataset}
\vspace{3mm}
\end{subfigure}
\begin{subfigure}{\linewidth}
\centering
\begin{tcblisting}{mylistingbox}
Q: "What kind of scene is shown?"
# 'what kind of scene' -> Scene
\end{tcblisting}
\vspace{-2mm}
\caption{Example of \textsc{Scene} problem in EarthVQA dataset}
\vspace{3mm}
\end{subfigure}
\begin{subfigure}{\linewidth}
\centering
\begin{tcblisting}{mylistingbox}
Q: "How many cargo ships are visible in the image?"
# 'how many'  -> Quantity
# 'cargo ship' (object class) -> Category
\end{tcblisting}
\vspace{-2mm}
\caption{Example of \textsc{Quantity}\,+\,\textsc{Category} problem in GEOBench-VLM dataset}
\end{subfigure}
}
\vspace{-2mm}
\caption{Examples of rule-based mapping of natural language problems.}
\label{fig:rule-examples-natural}
\end{figure}

\begin{figure}[h]
\centering
{
\tcbset{
    mylistingbox/.style={
        colback=white,
        colframe=black!15,
        boxrule=0.6pt,
        arc=2pt,
        left=6pt, right=6pt, top=5pt, bottom=5pt,
        boxsep=0pt,
        fontupper=\small,
        listing only,
        listing options={
            basicstyle=\ttfamily\small,
            breaklines=true,
            columns=fullflexible,
            aboveskip=0pt,
            belowskip=0pt
        }
    }
}

\begin{subfigure}{\linewidth}
\centering
\begin{tcblisting}{mylistingbox}
def f(image, a):
    x1, x2 = a  # x = ("water", "vegetation")
    return x1 if np.sum(segment(image, x1)) > np.sum(segment(image, x2)) else x2
# np.sum(.)        -> Coverage
# > with selection -> Comparison
# "water", ...     -> Scene
\end{tcblisting}
\vspace{-2mm}
\caption{Example of \textsc{Coverage}\,+\,\textsc{Comparison}\,+\,\textsc{Scene} problem generated in \pname{} self-play}
\vspace{3mm}
\end{subfigure}
\begin{subfigure}{\linewidth}
\centering
\begin{tcblisting}{mylistingbox}
def f(image, a):
    x1, x2 = a  # ("vehicle", "road")
    m2 = segment(image, x2)
    dt = distance_transform_edt(~m2)
    m1 = segment(image, x1)
    i = int(np.argmin([dt[ndi.center_of_mass(m)] for m in m1]))
    cy, cx = ndi.center_of_mass(m1[i])
    return ("TL" if cx < W/2 and cy < H/2 else
            "TR" if cy < H/2              else
            "BL" if cx < W/2              else "BR")
# "vehicle"                              -> Category
# distance_transform_edt, center_of_mass -> Relation
# "TL", "TR", ...                        -> Direction
\end{tcblisting}
\vspace{-2mm}
\caption{Example of \textsc{Category}\,+\,\textsc{Relation}\,+\,\textsc{Direction} problem generated in \pname{} self-play}
\end{subfigure}
}
\vspace{-2mm}
\caption{Examples of rule-based mapping of problems generated in \pname{} self-play.}
\label{fig:rule-examples-geozero}
\end{figure}

\clearpage
\newpage

\section{Qualitative analysis of the three reasoning modes}
\label{app:qualitative-analysis-three-modes}

This section supplements the qualitative analysis presented in Section~\ref{sec:discussion} (Figure~\ref{fig:qual-deduction}).
We extend the single deduction example to all three reasoning modes, with one representative example per mode drawn from a distinct image and program.

\begin{figure}[h]
    \centering
    \includegraphics[width=\linewidth]{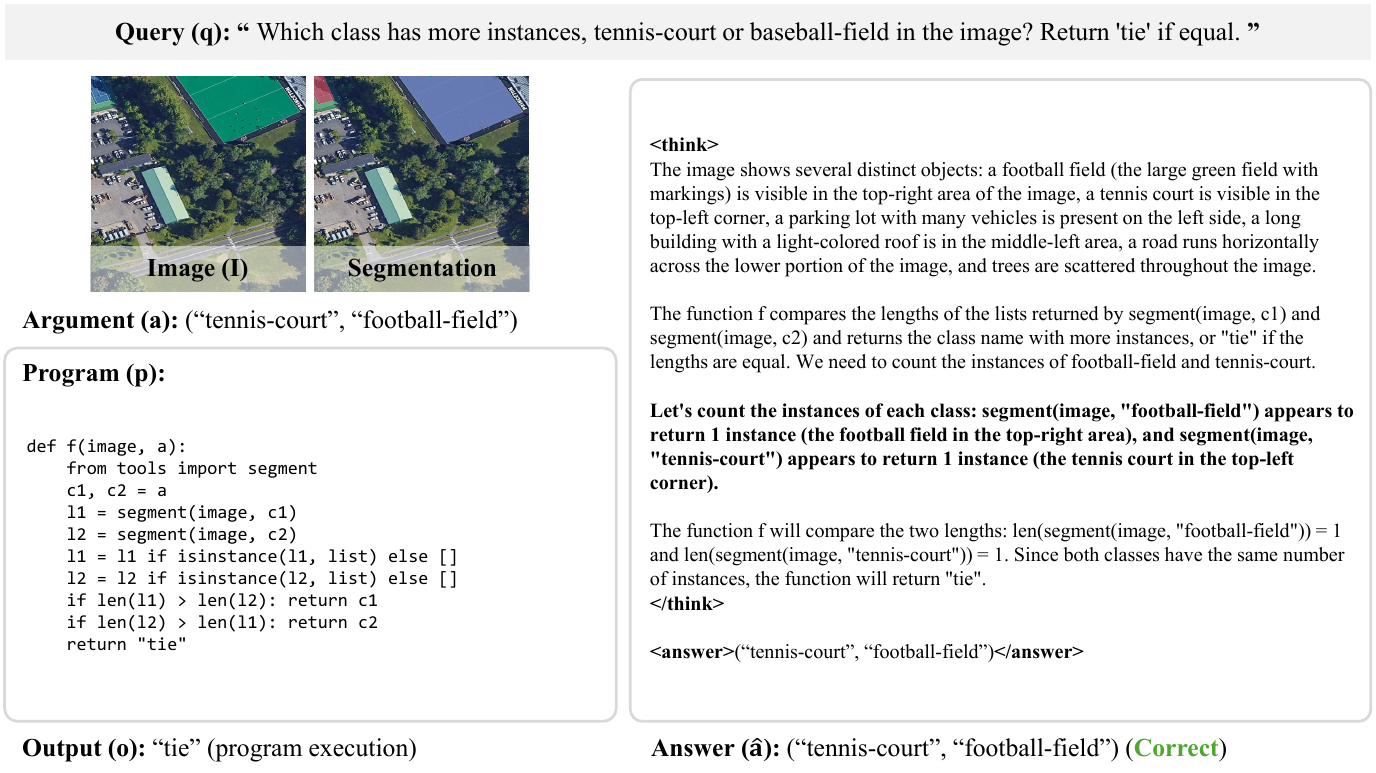}
    \caption{%
        \textbf{Qualitative analysis of abduction.}
        Given $(I, p, o)$, the solver infers $\hat{a}$ by simulating $p$ forward and searching for arguments whose execution reproduces the observed output $o$.%
    }
    \label{fig:qual-abduction}
\end{figure}

\begin{figure}[h]
    \centering
    \includegraphics[width=\linewidth]{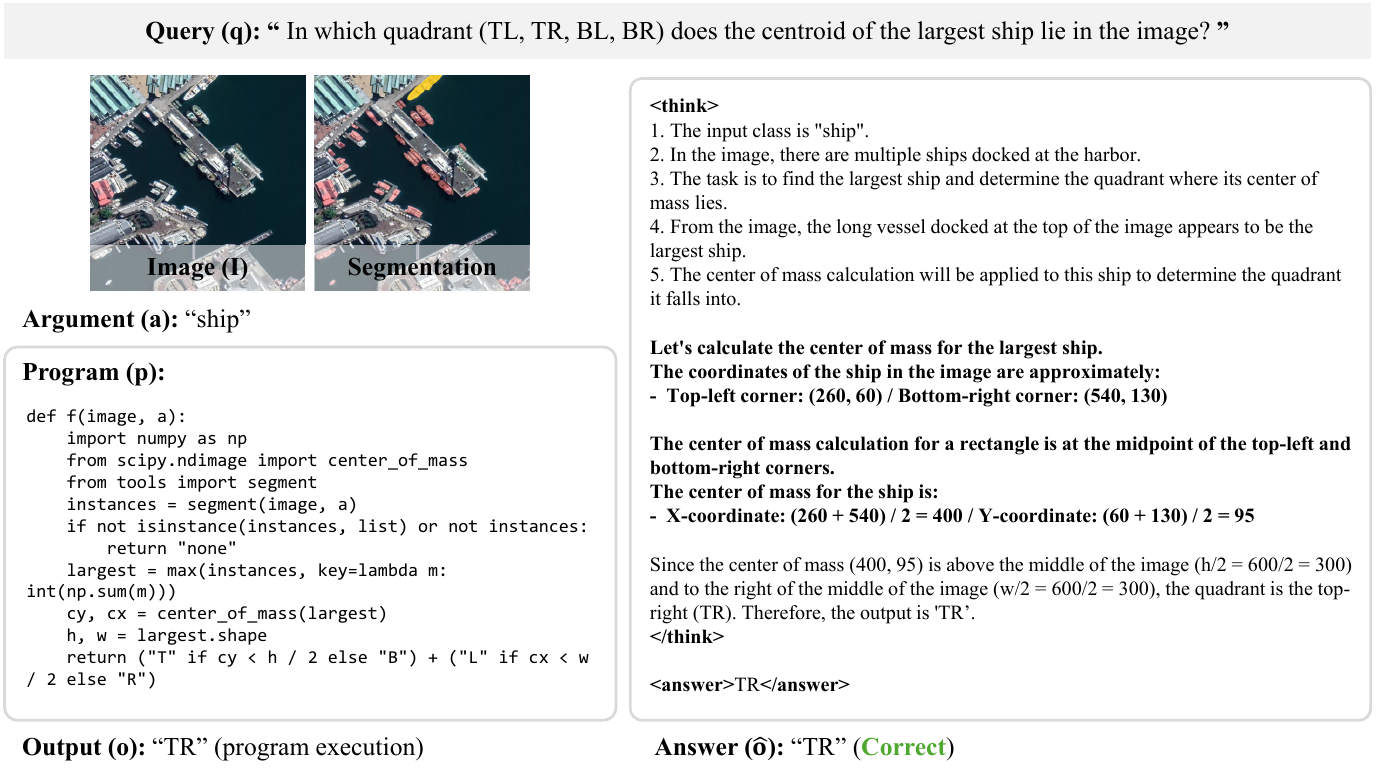}
    \caption{%
        \textbf{Qualitative analysis of deduction (reproduced from Figure~\ref{fig:qual-deduction}).}
        Given $(I, p, a)$, the solver follows the control flow of $p$ and predicts $\hat{o}$, which exactly matches the program-executed label $o = p(a; I)$.%
    }
    \label{fig:qual-deduction-app}
\end{figure}

\begin{figure}[h]
    \centering
    \includegraphics[width=\linewidth]{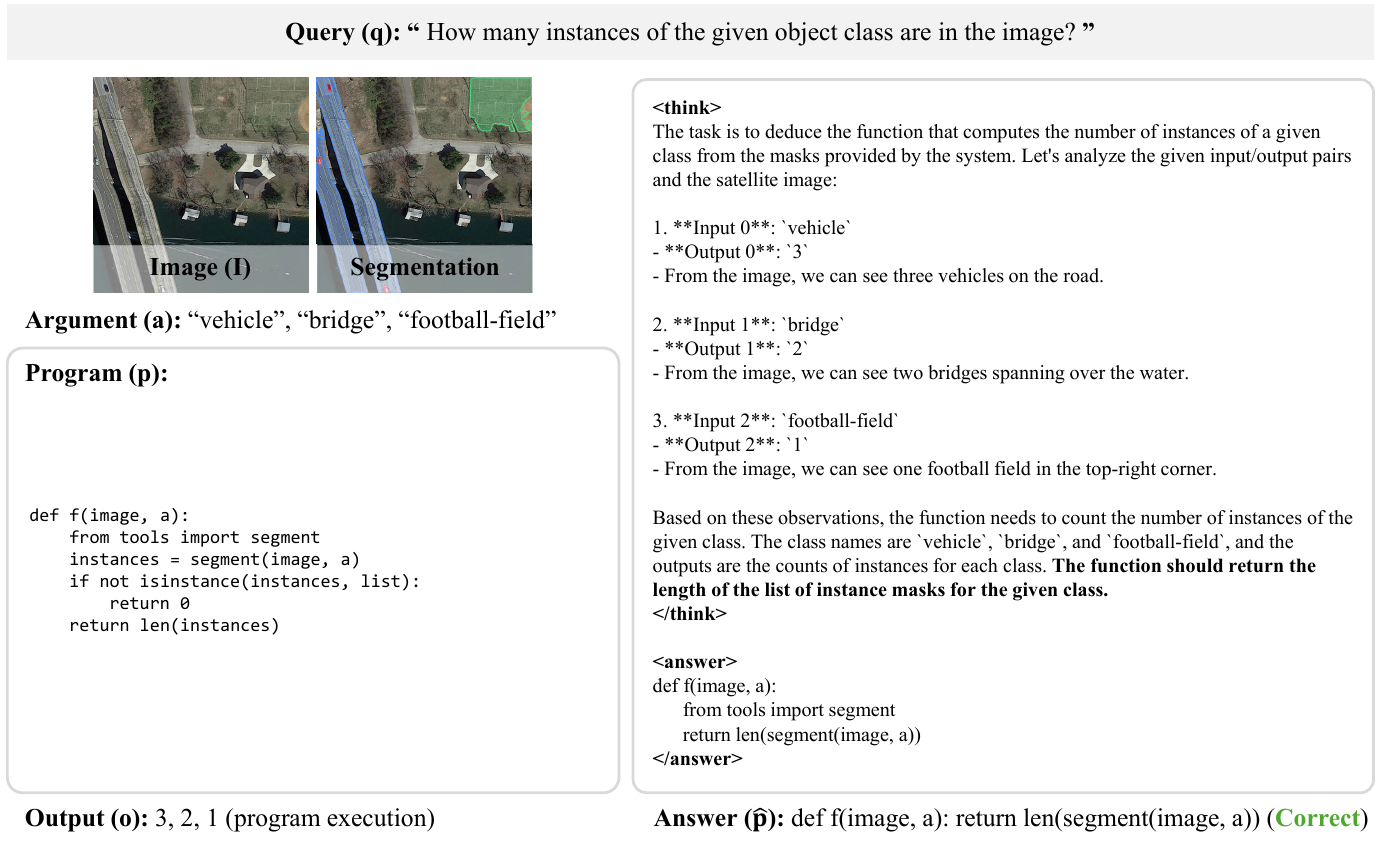}
    \caption{%
        \textbf{Qualitative analysis of induction.}
        Given input-output pairs $\{(a_t, o_t)\}_{t \in \mathcal{V}}$, the solver synthesizes a program $\hat{p}$ consistent with the visible pairs.%
    }
    \label{fig:qual-induction}
\end{figure}

For each mode, we present the visible elements of the triplet $(p, a, o)$, the solver's Chain-of-Thought, and its prediction verified against the program-executed output.
In \textbf{abduction} (Figure~\ref{fig:qual-abduction}), given $(I, p, o)$, the solver simulates $p$ forward and searches for an argument $\hat{a}$ whose execution reproduces $o$.
In \textbf{deduction} (Figure~\ref{fig:qual-deduction-app}), given $(I, p, a)$, the solver follows the control flow of $p$ and predicts $\hat{o}$; this example is reproduced from the main paper for direct comparison.
In \textbf{induction} (Figure~\ref{fig:qual-induction}), given input-output pairs $\{(a_t, o_t)\}_{t \in \mathcal{V}}$, the solver synthesizes $\hat{p}$ consistent with the visible pairs.
Across all modes, the solver's reasoning is anchored in concrete image regions rather than treating $p$ as an opaque label generator, indicating that self-play encourages the program to act as a procedural scaffold grounded in image evidence.

\clearpage
\newpage

%%%%%%%%%%%%%%%%%%%%%%%%%%%%%%%%%%%%%%%%%%%%%%% (Discussion 3: Done)

\section{Primitives library and usage in self-play}
\label{app:primitives}

This section supplements the analysis presented in Section~\ref{sec:discussion} (Figure~\ref{fig:operator-usage}).
We describe (i) the primitive library $\mathcal{F}$ exposed by the call interface, (ii) the procedure for detecting primitive invocations in programs, and (iii) the resulting per-primitive usage statistics.

\subsection{Primitive library}
\label{app:primitive-library}

The call interface $\mathcal{F}$ exposes 22 primitives organized into three groups.
The first two groups follow the standard geometric-topological distinction in spatial information systems~\citep{egenhofer1991point}; we extend it with a third
group for operations over sets of masks.
\emph{Geometric primitives} extract metric attributes from a single mask, such as area, bounding box, centroid, or principal orientation.
\emph{Topological primitives} capture spatial relations between two masks (adjacency, containment, distance, overlap, proximity, relative direction), or between a mask and image partitions such as grids and quadrants.
\emph{Aggregation primitives} reduce or transform over a set of masks, producing outputs such as a count, an extremum, a filtered subset, a presence check, or a union.

Table~\ref{tab:primitive-library} lists the primitives in each
group.

\begin{table}[h]
\centering
\caption{%
    \textbf{Primitive library $\mathcal{F}$.}
    Within each group, primitives are listed alphabetically.%
}
\label{tab:primitive-library}
\footnotesize
\setlength{\tabcolsep}{8pt}
\renewcommand{\arraystretch}{1.3}
\begin{tabular}{@{}l p{10.5cm}@{}}
\toprule
\textbf{Group} & \textbf{Primitives} \\
\midrule
\textit{Geometric} &
\texttt{area}, \texttt{bbox}, \texttt{centroid},
\texttt{orientation} \\
\textit{Topological} &
\texttt{adjacent}, \texttt{contains}, \texttt{distance},
\texttt{grid}, \texttt{in\_cell}, \texttt{nearest},
\texttt{overlaps}, \texttt{quadrant}, \texttt{relpos} \\
\textit{Aggregation} &
\texttt{argmin/argmax}, \texttt{components}, \texttt{count},
\texttt{exists}, \texttt{extreme}, \texttt{filter\_by},
\texttt{largest/smallest}, \texttt{mean\_position},
\texttt{union} \\
\bottomrule
\end{tabular}
\end{table}

\subsection{Deterministic detection rules}
\label{app:primitive-rules}

For each program $p$ constructed by the proposer, we determine its set of primitives $\mathcal{A}(p) \subseteq \mathcal{F}$ via anchored regular-expression matching on the source code.
A primitive is added to $\mathcal{A}(p)$ if any of its associated patterns matches a substring of $p$.
Patterns are designed to identify the operations a program performs, regardless of surface syntax, covering both explicit calls to a primitive (e.g., \texttt{centroid(m)}) and the standard library idioms that realize the same operation inline (e.g., \texttt{ndi.center\_of\_mass(m)}).

\paragraph{Pattern construction.}
We derived these patterns in two stages.
First, for each primitive, we enumerated the standard library functions and idioms that implement it (e.g., \texttt{np.any} for \texttt{exists}, \texttt{scipy.ndimage.center\_of\_mass} for \texttt{centroid}).
Second, we inspected a sample of programs and added patterns for any recurring idioms not covered by the initial set.
The full set of patterns and their match counts are reported alongside each primitive in Table~\ref{tab:primitive-evidence}.
%
% These counts also serve as implicit validation: patterns with high match counts and group-consistent semantics, i.e., consistent meaning across the patterns grouped under the same primitive, correspond to genuine primitive invocations rather than spurious matches.

\paragraph{Verification.}
We verified the matching procedure by manually annotating a held-out sample of 100 programs and comparing the predicted
primitive set $\mathcal{A}(p)$ against expert labels, observing every problem agreed on primitive membership.

% \subsection{Deterministic Detection Rules}
% \label{app:primitive-rules}

% For each program $p$ constructed by the proposer, we determine its set of primitives
% $\mathcal{A}(p) \subseteq \mathcal{F}$ via anchored regular-expression matching on the source code.
% %
% A primitive is added to $\mathcal{A}(p)$ if any of its associated patterns matches
% a substring of $p$.
% %
% Patterns include both explicit calls (e.g., \texttt{centroid(m)}) and idiomatic inline usage (e.g., \texttt{ndi.center\_of\_mass(m)}), crediting programs for the operations they perform regardless of surface syntax.
% %
% The full set of patterns and their match counts are given alongside each primitive in Tables~\ref{tab:primitive-evidence} and~\ref{tab:primitive-evidence-tail}.

\subsection{Usage statistics}
\label{app:primitive-usage}

Across the 6{,}558 programs, Table~\ref{tab:primitive-evidence} shows the program count, percentage, and top detection patterns for each primitive. 
Every group contains primitives appearing in over 1{,}000 programs, with \texttt{centroid} (55.4\%), \texttt{exists} (48.7\%), \texttt{quadrant} (32.9\%), \texttt{extreme} (28.8\%), \texttt{count} (25.7\%), and \texttt{area} (25.0\%) dominating the distribution.
%
% The long tail of rarely used primitives is itself a finding: their availability is a precondition for use, and their sparse occurrence reveals the spatial concepts the proposer chooses to exercise during self-play.

\begin{table}[t]
\centering
\caption{%
    \textbf{Per-primitive usage across 6{,}558 programs.}
    A program is counted once per primitive. Pattern hits in
    parentheses give the number of programs matched by each pattern;
    a program may match multiple patterns of the same primitive.
    Within each group, primitives are ordered by frequency.%
}
\label{tab:primitive-evidence}
\footnotesize
\setlength{\tabcolsep}{6pt}
\renewcommand{\arraystretch}{1.25}
\begin{tabular}{@{}l l r r p{6.0cm}@{}}
\toprule
\textbf{Group} & \textbf{Primitive} &
\textbf{Count} & \textbf{\%} &
\textbf{Top detection patterns} \\
\midrule
\multirow{4}{*}{\textit{Geometric}}
 & \texttt{centroid}    & 3{,}630 & 55.4 &
\texttt{center\_of\_mass} (3{,}558), \texttt{cx, cy} (2{,}331) \\
 & \texttt{area}        & 1{,}642 & 25.0 &
\texttt{np.sum} (1{,}604), identifier \texttt{area} (41) \\
 & \texttt{bbox}        &    282 &  4.3 &
\texttt{np.where} (255), \texttt{xs/ys.min/max} (136) \\
 & \texttt{orientation} &      3 &  0.0 &
\texttt{regionprops} (3) \\
\midrule
\multirow{9}{*}{\textit{Topological}}
 & \texttt{quadrant} & 2{,}157 & 32.9 &
\texttt{cy < H/2} (2{,}002), \texttt{cx < W/2} (1{,}838) \\
 & \texttt{nearest}  &    741 & 11.3 &
\texttt{distance\_transform\_edt} (713),
\texttt{min\_dist}, \texttt{nearest} (382) \\
 & \texttt{distance} &    272 &  4.1 &
Euclidean expansion (249), \texttt{np.linalg.norm} (2) \\
 & \texttt{grid}     &     75 &  1.1 &
\texttt{H/N}, \texttt{W/N} for $N \geq 3$ (68) \\
 & \texttt{relpos}   &     70 &  1.1 &
cardinal literals (\texttt{"N"}, \texttt{"NE"}, \dots) (70) \\
 & \texttt{overlaps} &     52 &  0.8 &
\texttt{np.any(\,\&\,)} (39), \texttt{combined \&} (23) \\
 & \texttt{adjacent} &      3 &  0.0 &
\texttt{adjacent} (3) \\
 & \texttt{contains} &      1 &  0.0 &
\texttt{contains} (1) \\
 & \texttt{in\_cell} &      0 &  0.0 &
--- \\
\midrule
\multirow{9}{*}{\textit{Aggregation}}
 & \texttt{exists}           & 3{,}194 & 48.7 &
\texttt{np.any} (3{,}189), \texttt{.any()} (9) \\
 & \texttt{extreme}          & 1{,}887 & 28.8 &
quadrant literals (1{,}050), direction literals (841) \\
 & \texttt{count}            & 1{,}686 & 25.7 &
\texttt{len(masks[}\dots\texttt{])} (1{,}168),
\texttt{len(instances)} (518) \\
 & \texttt{argmin/argmax}    & 1{,}129 & 17.2 &
\texttt{argmax} (1{,}079), \texttt{argmin} (70) \\
 & \texttt{union}            &    985 & 15.0 &
\texttt{|=} (977), \texttt{combined =} (960) \\
 & \texttt{largest/smallest} &    253 &  3.9 &
\texttt{max(\dots, key=)} (220), \texttt{sorted} (29) \\
 & \texttt{mean\_position}   &    152 &  2.3 &
\texttt{avg\_y} (115), \texttt{avg\_x} (113) \\
 & \texttt{filter\_by}       &     11 &  0.2 &
list comprehension with \texttt{if} (11) \\
 & \texttt{components}       &      1 &  0.0 &
\texttt{from scipy.ndimage import label} (1) \\
\bottomrule
\end{tabular}
\end{table}

\clearpage
\newpage

\section{Full results}
\label{app:full-results}

This section reports the experimental results in
Section~\ref{sec:exp}. We provide (i) per-subtask breakdowns of
the main VQA results in Table~\ref{tab:main-results}
(Section~\ref{app:main-subtask}), (ii) per-subtask breakdowns of
the ablation study in Table~\ref{tab:ablation-main}
(Section~\ref{app:ablation-subtask}). 

%and (iii) zero-shot evaluation of \pname{} on the auxiliary scene classification and counting benchmarks introduced in Sections~\ref{app:datasets-scene}--\ref{app:datasets-counting}(Section~\ref{app:scene-counting-eval}).

\subsection{Full sub-task results on VQA benchmarks}
\label{app:main-subtask}

Table~\ref{tab:main-results} in the main paper presents the performance of \pname{} and other baselines on RSVQA-HR and EarthVQA at the per-subtask level, and on GEOBench-VLM at the category level by averaging the subtasks within each category.
Here, we provide the per-subtask breakdown for all benchmark datasets, including GEOBench-VLM across all 17 spatial-understanding tasks.

\colorlet{curatedcol}{gray!12}
\colorlet{gaincol}{cyan!10}

\newcolumntype{M}{>{\centering\arraybackslash}p{1.15cm}}
\newcolumntype{C}{>{\columncolor{curatedcol}\centering\arraybackslash}p{1.15cm}}
\newcolumntype{O}{>{\centering\arraybackslash}p{1.15cm}}

\begin{table}[h]
\centering
\footnotesize
\renewcommand{\arraystretch}{1.2}

\caption{
\textbf{Full results on VQA benchmarks.}
Performance of \pname{}, instantiated from two base VLMs (LLaVA and Qwen), is compared against six baselines on RSVQA-HR, EarthVQA, and GEOBench-VLM.
For each model, the number of human-curated QA pairs used for fine-tuning is listed; baseline models fine-tuned on fully curated data are grey shaded.
The best result in each row is shown in \textbf{bold}, and our results are highlighted in blue when they surpass the corresponding base (LLaVA or Qwen).
}

\begin{adjustbox}{width=0.95\textwidth}
\begin{tabular}{@{}l M M C C C C O O@{}}
\toprule
& \multicolumn{2}{c}{\textit{Base}}
& \multicolumn{4}{c}{\textit{Conventional}}
& \multicolumn{2}{c}{\textit{Self-play}} \\
\cmidrule(lr){2-3} \cmidrule(lr){4-7} \cmidrule(lr){8-9}
\textbf{Benchmark / Task}
  & LLaVA & Qwen
  & GeoChat & VHM & EarthDial & RSThinker
  & \textbf{\pname{}\textsubscript{L}} & \textbf{\pname{}\textsubscript{Q}} \\
\midrule
\textit{\# Curated}
  & 0 & 0 & 318K & 1.4M & 11.1M & 380K & 0 & 0 \\
\midrule

\multicolumn{9}{@{}l}{\textit{RSVQA-HR}} \\
\addlinespace[2pt]
~~Presence                         & 57.9 & 60.1 & 57.9 & \textbf{72.3} & 60.0 & 55.4 & \cellcolor{gaincol}61.9 & \cellcolor{gaincol}63.2 \\
~~Count                            & 30.3 & 35.2 & 25.5 & 13.5 & 25.9 & 22.1 & \cellcolor{gaincol}30.5 & \cellcolor{gaincol}\textbf{36.6} \\
~~Area                             &  2.7 & 10.3 & \textbf{41.6} & 13.5 & 38.8 & 22.1 & \cellcolor{gaincol}40.3 & \cellcolor{gaincol}16.5 \\
~~Comparison                       & 52.5 & 69.7 & 74.8 & 70.7 & \textbf{78.9} & 68.7 & \cellcolor{gaincol}57.5 & \cellcolor{gaincol}72.0 \\
\midrule
~~\textbf{Average}                 & 35.9 & 43.8 & 50.0 & 42.5 & \textbf{50.9} & 42.1 & \cellcolor{gaincol}47.6 & \cellcolor{gaincol}47.1 \\
\midrule

\multicolumn{9}{@{}l}{\textit{EarthVQA}} \\
\addlinespace[2pt]
~~Basic Judging                    & 78.6 & 78.0 & 66.6 & 79.0 & 76.2 & 69.4 & \cellcolor{gaincol}\textbf{82.2} & \cellcolor{gaincol}78.2 \\
~~Reasoning-based Judging          & 55.8 & \textbf{73.0} & 37.0 & 71.8 & 41.4 & 42.6 & 53.8 & 68.6$^\dagger$ \\
~~Reasoning-based Counting         & 24.8 & 23.4 & 16.6 & 12.8 &  5.6 & 15.4 & \cellcolor{gaincol}\textbf{33.8} & \cellcolor{gaincol}26.6 \\
~~Basic Counting                   & 60.6 & 62.4 & 26.0 & 39.6 & 53.0 & 45.8 & \cellcolor{gaincol}\textbf{70.2} & \cellcolor{gaincol}69.4 \\
~~Object Situation Analysis        & 28.2 & 29.4 &  2.2 & 23.2 &  4.0 & 25.2 & \cellcolor{gaincol}\textbf{30.4} & \cellcolor{gaincol}30.2 \\
~~Comprehensive Analysis           & 19.6 & 33.6 &  7.8 & 24.6 & 11.0 & 19.0 & 19.5 & \cellcolor{gaincol}\textbf{34.4} \\
\midrule
~~\textbf{Average}                 & 44.6 & 50.0 & 26.0 & 41.8 & 31.9 & 36.2 & \cellcolor{gaincol}48.3 & \cellcolor{gaincol}\textbf{51.2} \\
\midrule

\multicolumn{9}{@{}l}{\textit{GEOBench-VLM}} \\
\addlinespace[2pt]
~~Spatial Relation Classification  & 35.2 & 68.2 & 53.4 & 57.1 & 48.9 & 62.9 & \cellcolor{gaincol}36.3 & \cellcolor{gaincol}\textbf{70.0} \\
~~Building Counting                & 17.6 & 29.4 & 17.6 & 19.4 & 21.2 & 12.9 & \cellcolor{gaincol}21.2 & \cellcolor{gaincol}\textbf{37.1} \\
~~General Vehicle Counting         & 12.7 & \textbf{38.7} & 16.0 & 16.7 & 16.0 & 28.0 & \cellcolor{gaincol}13.3 & \textbf{38.7} \\
~~Specific Vehicle Counting        & 17.0 & 29.0 & 17.4 & 21.0 & 23.2 & 12.9 & \cellcolor{gaincol}19.2 & \cellcolor{gaincol}\textbf{33.9} \\
~~General Aircraft Counting       & 18.0 & 32.0 & 22.0 & 14.0 & 21.0 & 29.0 & \cellcolor{gaincol}22.0 & \cellcolor{gaincol}\textbf{33.0} \\
~~Specific Aircraft Counting      & 22.1 & 22.9 & 16.4 & 20.7 & 22.9 & 21.4 & 20.7 & \cellcolor{gaincol}\textbf{27.9} \\
~~Marine Debris Counting           & 17.0 & 35.0 & 18.0 & 29.0 & 26.0 & 37.0 & 16.0 & \cellcolor{gaincol}\textbf{38.0} \\
~~Trees Counting                   & 15.3 & \textbf{29.4} & 21.2 & 24.7 & 21.2 &  5.9 & \cellcolor{gaincol}20.0 & 28.2 \\
~~Tree Health Assessment           & 17.8 & \textbf{24.8} & 19.1 & 19.1 & 16.6 &  8.9 & 15.9 & 19.1 \\
~~Water Bodies Counting            & 32.9 & 63.5 & 21.2 & 48.2 & 30.6 & 28.2 & \cellcolor{gaincol}37.7 & \cellcolor{gaincol}\textbf{69.4} \\
~~Scene Classification             & 67.7 & 79.5 & 79.5 & 83.6 & \textbf{94.0} & 88.4 & \cellcolor{gaincol}68.0 & 79.3 \\
~~Land Use Classification          & 46.6 & 53.9 & 51.0 & 60.0 & \textbf{62.0} & 49.5 & 46.1 & \cellcolor{gaincol}55.1 \\
~~Crop Type Classification         & 25.5 & \textbf{30.9} & 18.2 & 21.8 & 25.5 & 10.9 & \cellcolor{gaincol}27.3 & 27.3 \\
~~Aircraft Type Classification     & 40.0 & 48.0 & 36.0 & 48.0 & 36.0 & 45.0 & \cellcolor{gaincol}42.0 & \cellcolor{gaincol}\textbf{53.0} \\
~~Ship Type Classification         & 30.4 & 39.6 & 32.9 & 24.6 & 35.7 & 27.5 & \cellcolor{gaincol}30.9 & \cellcolor{gaincol}\textbf{42.0} \\
~~Disaster Type Classification     & 11.8 & \textbf{70.6} & 58.8 & 64.7 & 17.6 & 23.5 & 11.8 & 64.7 \\
~~Fire Risk Assessment             & 17.5 & 20.0 & 16.7 & \textbf{29.2} & 15.8 & 18.3 & 16.7 & 20.0 \\
\midrule
~~\textbf{Average}                 & 26.2 & 42.1 & 30.3 & 35.4 & 31.4 & 30.0 & \cellcolor{gaincol}27.4 & \cellcolor{gaincol}\textbf{43.3} \\
\bottomrule
\end{tabular}
\end{adjustbox}

\label{tab:main-results-full}
\end{table}

\clearpage
\newpage

\subsection{Full sub-task results on the ablation study}
\label{app:ablation-subtask}

Table~\ref{tab:ablation-main} in the main paper summarizes the ablation across the four task categories of GEOBench-VLM.
Here, we provide the per-subtask breakdown for each variant of \pname{}.
\emph{Singleton} variants keep one solver mode active (A: Abduction, D: Deduction, I: Induction); \emph{Drop-one} variants disable one of the three; \emph{Training} variants (\emph{BaseGen} and \emph{SolvOnly}) retain all three modes but modify problem generation or proposer reward as described in Section~\ref{sec:exp:ablation}; \emph{Full} denotes the default configuration.

\begin{table}[h]
\centering
\caption{\textbf{Full ablation study results on GEOBench-VLM.}
Columns specify which solver modes are active: A (Abduction), D (Deduction), I (Induction). 
\emph{Singleton Variants} keeps one mode;
\emph{Drop-one Variants} removes one;
\emph{Training Variants} use all three reasoning modes.
\textit{Full} denotes our default configuration.
The best result in each row is shown in \textbf{bold} and the second-best is shown in \underline{underline}.}
\label{tab:ablation-pertask}
\scriptsize
\setlength{\tabcolsep}{4pt}
\renewcommand{\arraystretch}{1.2}
\newcolumntype{N}{>{\centering\arraybackslash}p{3.4em}}
\resizebox{\textwidth}{!}{
\begin{tabular}{@{}l N N N N N N N N N@{}}
\toprule
& \multicolumn{3}{c}{\textit{Singleton Variants}}
& \multicolumn{3}{c}{\textit{Drop-one Variants}}
& \multicolumn{2}{c}{\textit{Training Variants}}
& \textit{Full} \\
\cmidrule(lr){2-4} \cmidrule(lr){5-7} \cmidrule(lr){8-9} \cmidrule(lr){10-10}
\textbf{Config.}
  & Abd & Ded & Ind
  & $-$Abd & $-$Ded & $-$Ind
  & BaseGen & SolvOnly
  & \pname{} \\
\textit{Active Modes}
  & A & D & I
  & D,I & A,I & D,A
  & D,A,I & D,A,I
  & D,A,I \\
\midrule
\multicolumn{10}{@{}l}{\textit{Object Localization \& Counting}} \\
~~Spatial Relation Classification & 67.8 & 67.8 & 67.8 & 67.8 & 67.8 & \underline{68.5} & 66.5 & 68.2 & \textbf{70.0} \\
~~Building Counting                & 33.5 & 35.3 & 32.9 & 32.9 & 35.3 & 35.9 & \underline{36.5} & 32.9 & \textbf{37.1} \\
~~General Vehicle Counting        & 34.7 & 36.7 & 36.0 & 34.7 & 33.3 & 36.7 & \underline{38.0} & 28.7 & \textbf{38.7} \\
~~Specific Vehicle Counting       & 27.7 & 28.6 & 27.7 & 26.3 & 30.8 & 30.4 & \underline{31.7} & 27.7 & \textbf{33.9} \\
~~General Aircraft Counting       & 29.0 & \textbf{33.0} & 28.0 & 28.0 & 30.0 & 31.0 & 30.0 & 29.0 & \textbf{33.0} \\
~~Specific Aircraft Counting      & 22.9 & 25.0 & \textbf{28.6} & 22.1 & 25.7 & 24.3 & 21.4 & 22.1 & \underline{27.9} \\
~~Marine Debris Counting          & 33.0 & 35.0 & 34.0 & \textbf{39.0} & 33.0 & 35.0 & 32.0 & 36.0 & \underline{38.0} \\
~~Trees Counting                   & 22.4 & 25.9 & 24.7 & \textbf{29.4} & 24.7 & 24.7 & 27.1 & 23.5 & \underline{28.2} \\
~~Tree Health Assessment          & \underline{21.7} & 19.1 & 19.7 & 19.7 & 19.1 & 17.8 & \textbf{24.8} & 14.6 & 19.1 \\
~~Water Bodies Counting           & 62.4 & 65.9 & 64.7 & 64.7 & 64.7 & \textbf{70.6} & 65.9 & 60.0 & \underline{69.4} \\
\midrule
~~Average                        & 35.5 & 37.2 & 36.4 & 36.5 & 36.4 & \underline{37.5} & 37.4 & 34.3 & \textbf{39.5} \\
\midrule
\multicolumn{10}{@{}l}{\textit{Scene Understanding}} \\
~~Scene Classification             & 78.1 & \underline{78.6} & 76.6 & 75.4 & 76.9 & 78.1 & \underline{78.6} & 75.4 & \textbf{79.3} \\
~~Land Use Classification          & 54.4 & 54.9 & 53.4 & 52.5 & 53.7 & 54.7 & \textbf{55.4} & 53.2 & \underline{55.1} \\
~~Crop Type Classification        & 21.8 & \underline{23.6} & 18.2 & 16.4 & 21.8 & \underline{23.6} & 20.0 & 21.8 & \textbf{27.3} \\
\midrule
~~Average                        & 51.4 & \underline{52.4} & 49.4 & 48.1 & 50.8 & 52.1 & 51.3 & 50.1 & \textbf{53.9} \\
\midrule
\multicolumn{10}{@{}l}{\textit{Object Classification}} \\
~~Ship Type Classification        & 40.6 & 40.6 & 41.5 & 38.2 & \underline{43.5} & 38.6 & \textbf{46.0} & 39.6 & 42.0 \\
~~Aircraft Type Classification    & 50.0 & 53.0 & 51.0 & 49.0 & \textbf{55.0} & 50.0 & 39.1 & 51.0 & \textbf{53.0} \\
\midrule
~~Average                        & 45.3 & 46.8 & 46.3 & 43.6 & \textbf{49.3} & 44.3 & 42.6 & 45.3 & \underline{47.5} \\
\midrule
\multicolumn{10}{@{}l}{\textit{Event Detection}} \\
~~Disaster Type Classification    & 58.8 & 64.7 & 58.8 & 58.8 & 64.7 & 64.7 & \textbf{70.6} & \textbf{70.6} & 64.7 \\
~~Fire Risk Assessment            & 19.2 & 20.0 & 21.7 & 21.7 & \underline{22.5} & 20.8 & \textbf{23.3} & \underline{22.5} & 20.0 \\
\midrule
~~Average                        & 39.0 & 42.4 & 40.3 & 40.3 & 43.6 & 42.8 & \textbf{47.0} & \underline{46.6} & 42.4 \\
\midrule
\textbf{Total Average}                        & 39.9 & \underline{41.6} & 40.3 & 39.8 & 41.3 & 41.5 & \underline{41.6} & 39.8 & \textbf{43.3} \\
\bottomrule
\end{tabular}
}
\end{table}

\clearpage
\newpage
% \input{sec/300_app_prompts}

% \clearpage
% \newpage

% \input{checklist.tex}

\end{document}